\documentclass[10pt,journal,compsoc]{IEEEtran}
\usepackage[T1]{fontenc}
\usepackage{graphicx}
\usepackage{amsmath}
\usepackage{amsthm}
\usepackage{amssymb}
\usepackage{booktabs}
\usepackage{colortbl}
\usepackage{pifont}
\usepackage[ruled,linesnumbered]{algorithm2e}

\SetCommentSty{mycommfont}
\usepackage[table]{xcolor} 
\definecolor{Gray}{gray}{0.93}
\newtheorem{theo}{Theorem}
\usepackage{wrapfig}
\definecolor{LightYellow}{rgb}{0.9954, 0.98665, 0.94745}
\definecolor{LightGreen}{rgb}{0.96,0.995,0.975} 
\definecolor{mydarkred}{rgb}{0.4,0,0}
\definecolor{mydarkgreen}{rgb}{0,0.6,0}
\definecolor{purpleCustom}{RGB}{96, 64, 151}  
\definecolor{redCustom}{RGB}{195, 69, 24}  
\colorlet{highlightcolor}{black}

\usepackage[breaklinks=true,
            colorlinks,
            linkcolor = mydarkred,
            urlcolor  = mydarkred, 
            citecolor = mydarkgreen,
            bookmarks = false]{hyperref}

\usepackage[labelformat=parens]{subcaption}
\usepackage{multirow,amsfonts,makecell,array}
\usepackage{enumitem}
\usepackage{threeparttable}
\usepackage{xcolor}
\usepackage{graphicx}
\usepackage[capitalize,noabbrev]{cleveref}
\crefname{algocf}{Algorithm}{Algorithms}
\Crefname{algocf}{Algorithm}{Algorithms}

\graphicspath{{./fig/}}
\usepackage{bbm}
\usepackage[switch]{lineno}

\newif\ifpaperlined
\paperlinedfalse

\renewcommand{\linelabel}[1]{}
\ifpaperlined
  \colorlet{highlightcolor2}{red!75!black}
\else
  \colorlet{highlightcolor2}{black}
\fi

\newtheorem{Theorem}{Theorem}
\newtheorem{Lemma}{Lemma}

\ifCLASSOPTIONcompsoc
  \usepackage[nocompress]{cite}
\else
  \usepackage{cite}
\fi

\ifCLASSINFOpdf
\else
\fi

\begin{document}

\title{Improving Generalizability and Undetectability for Targeted Adversarial Attacks on Multimodal Pre-trained Models}

\author{Zhifang~Zhang,
    Jiahan~Zhang,
    Shengjie~Zhou,
   Qi~Wei,
   Shuo~He,
   Feng~Liu,
   Lei~Feng
\IEEEcompsocitemizethanks{
\IEEEcompsocthanksitem Zhifang Zhang and Lei Feng are with the School of Computer Science and Engineering, Southeast University, China.
E-mail: zzfofficial@gmail.com, fenglei@seu.edu.cn.
\IEEEcompsocthanksitem Jiahan Zhang is with the Department of Computer Science, Johns Hopkins University, USA.
E-mail: jhanzhang01@gmail.com.
\IEEEcompsocthanksitem Shengjie Zhou is with the College of Computer Science, Chongqing University, China.
E-mail: zshengjie@cqu.edu.cn.
\IEEEcompsocthanksitem Shuo He and Qi Wei are with the College of Computing and Data Science, Nanyang Technological University, Singapore.
E-mail: shuo.he@ntu.edu.sg; qi.wei@ntu.edu.sg.
\IEEEcompsocthanksitem Feng Liu is with the School of Computing and Information Systems, University of Melbourne, Australia.
E-mail: feng.liu1@unimelb.edu.au.
\IEEEcompsocthanksitem Zhifang Zhang and Jiahan Zhang are co-first authors.
\IEEEcompsocthanksitem Corresponding author: Lei Feng.
}}

\IEEEtitleabstractindextext{
\begin{abstract}
  Multimodal pre-trained models (e.g., ImageBind), which align distinct data modalities into a shared embedding space, have shown remarkable success across downstream tasks. 
  However, their increasing adoption raises serious security concerns, especially regarding targeted adversarial attacks.
  In this paper, we show that existing targeted adversarial attacks on multimodal pre-trained models still have limitations in two aspects: \emph{generalizability} and \emph{undetectability}. 
  Specifically, the crafted targeted adversarial examples (AEs) exhibit limited generalization to partially known or semantically similar targets in cross-modal alignment tasks (i.e., limited generalizability) and can be easily detected by simple anomaly detection methods (i.e., limited undetectability).
  To address these limitations, we propose a novel method called \textbf{P}roxy \textbf{T}argeted \textbf{A}ttack (PTA), which leverages multiple source-modal and target-modal proxies to optimize targeted AEs, ensuring they remain evasive to defenses while aligning with multiple potential targets. 
  {\color{highlightcolor}We also provide theoretical analyses to characterize the embedding-space trade-off between generalizability and undetectability, which motivates the proposed optimization objective.}
  Furthermore, experimental results demonstrate that PTA can achieve a high success rate across various related targets and remain undetectable against multiple anomaly detection methods.
\end{abstract}

}

\markboth{IEEE TRANSACTIONS ON PATTERN ANALYSIS AND MACHINE INTELLIGENCE}%
{Shell \MakeLowercase{\textit{et al.}}: Bare Demo of IEEEtran.cls for Computer Society Journals}

\maketitle

\IEEEdisplaynontitleabstractindextext

\section{Introduction}\label{sec:introduction}
\IEEEPARstart{W}ith the rapid expansion of data availability, computational resources, and advancements in model architectures, multimodal pre-trained models (e.g., Imagebind \cite{girdhar2023imagebind}) have demonstrated remarkable success \cite{wang2023onepeace, su2023pandagpt, xing2024seeing, girdhar2023imagebind}, which typically leverage contrastive learning to align multiple modalities into a shared latent space.
As powerful multimodal encoders, these models have been widely employed as foundational building blocks integrated into high-level systems for various downstream applications, including creative content generation \cite{xing2024seeing, su2023pandagpt, huang2024dreamtime,li2023blip2a} and cross-modal tasks \cite{jiang2024visual, chi2024eva,lerner2024crossmodal}. 
However, the widespread adoption of these multimodal pre-trained models has introduced new security threats \cite{zhao2024evaluating,schulhoff2023ignore,fan2024unbridled}. 
One of the most serious threats is targeted adversarial attacks \cite{zhang2024adversarial, zhao2024evaluating}, which specifically exploit the shared embedding space of such encoders to degrade the performance of downstream cross-modal matching tasks.

\begin{figure*}
  \centering
  \includegraphics[width=1.0\linewidth]{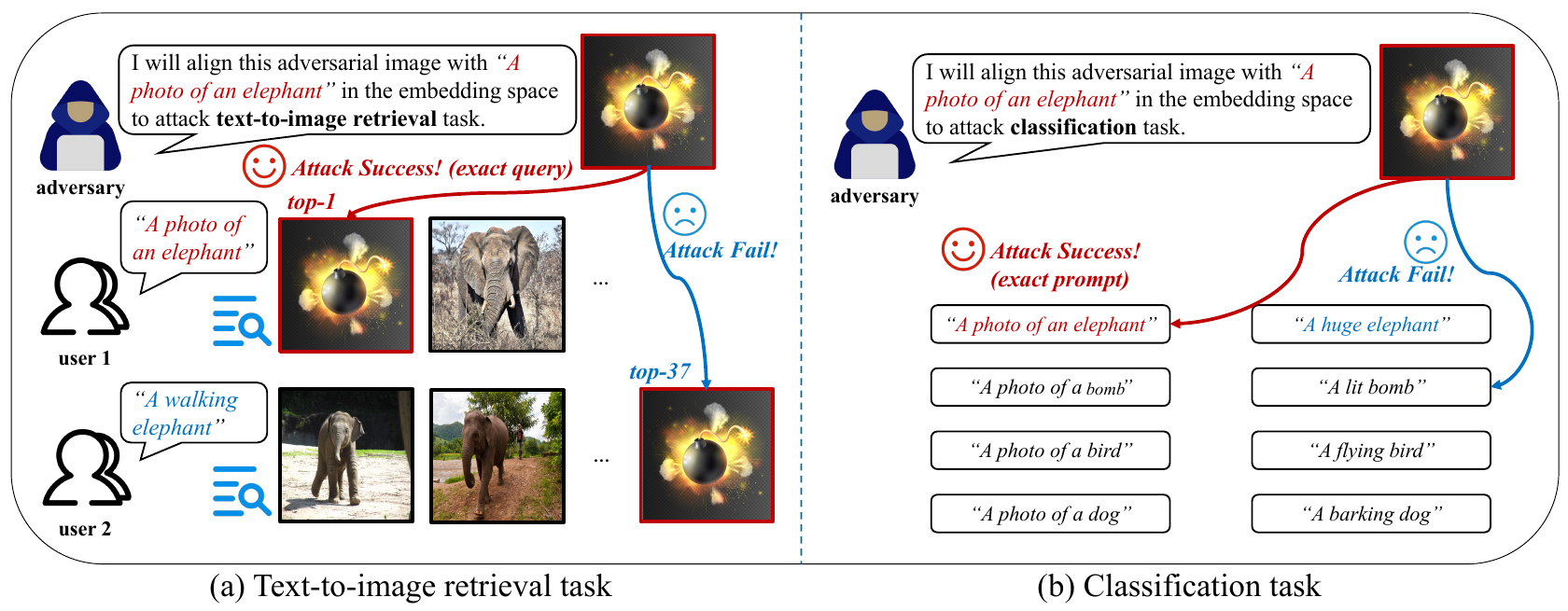}
  \caption{
  \textbf{Limited generalizability of the current targeted adversarial example (AE).}
    An AE crafted by Illusion Attack to align with ``A photo of an elephant'' \textbf{(a)} ranks top for that exact query in retrieval tasks but drops sharply on semantically similar queries like ``A walking elephant.'' Likewise, in classification \textbf{(b)}, it successfully fools the model with the exact prompt but fails on a slight variation (i.e., ``A huge elephant''), underscoring its poor generalization to unseen targets.
  }
  \vspace{-5mm}
  \label{fig:setting}
\end{figure*}

Previous work has focused on crafting targeted adversarial examples (AEs) by exploiting the shared embedding space of multimodal models to maximize the cosine similarity between each AE and its intended target embedding, which can be cross-modal \cite{zhang2024adversarial, zhao2024evaluating, inkawhich2023adversarialattacksfoundationalvision} or same-modal \cite{zhao2024evaluating, dou2024adversarialattacksmultimodalmodels}.
While optimizing the AE towards a same-modal target experimentally results in poor performance in cross-modal tasks, methods adopting a cross-modal target\footnote{In this paper, we refer to such methods \cite{zhang2024adversarial} as \textit{Illusion Attacks}.} can achieve a high attack success rate under ideal conditions.
However, we notice that the so-called Illusion Attacks suffer a sharp performance drop when tested on unseen targets, limiting their practical applicability.
For example, as shown in \cref{fig:setting}(a), in the text-to-image retrieval task, an AE crafted for the query ``A photo of an elephant'' ranks first for that exact query but significantly drops in ranking with the similar query of ``A walking elephant.''
Similarly, \cref{fig:setting}(b) shows that in classification tasks, the AE successfully fools the model with the identical prompt but performs poorly when slight variations like ``A huge elephant'' are introduced.
Notably, in real-world scenarios, adversaries typically possess only partial knowledge of the user’s input (e.g., relevant keywords or semantically similar examples) rather than the exact information.
Thus, targeted AEs must generalize beyond a single target to be effective in practice.
Moreover, such generalizable targeted AEs used as poison samples in the gallery of multimodal retrieval systems can inflict greater harm, as the ability to match a broader range of potential targets allows them to degrade system performance more effectively with fewer injected samples compared with untargeted AEs or conventional targeted AEs.
To summarize, improving the generalizability of targeted AEs (i.e., the ability to generalize to partially known or semantically similar targets) is essential for successful and impactful cross-modal alignment attacks.

\begin{figure*}[t]
  \centering 
  \includegraphics[width=1.0\linewidth]{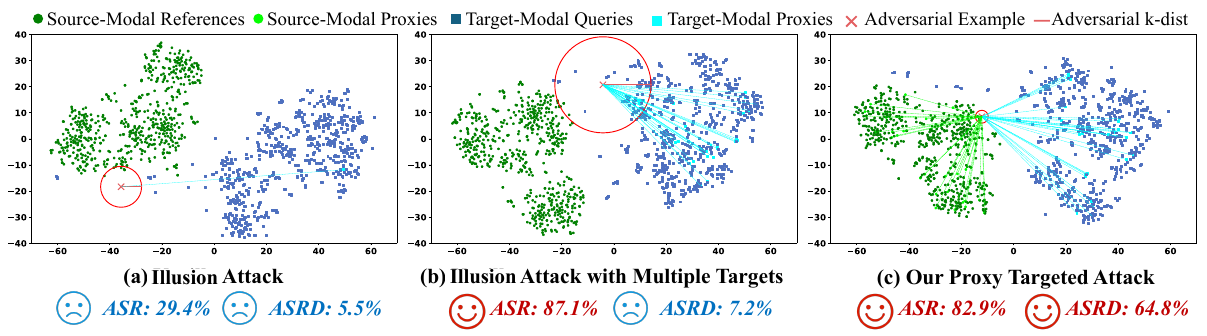}
  \caption{
  \textbf{t-SNE visualization of embedding space for three targeted attack strategies.}
  \textbf{(a)}~Illusion Attack optimized with a single target \cite{zhang2024adversarial}: low attack success rate (ASR) with different but semantically similar targets, and low ASR after anomaly detection (ASRD).
  \textbf{(b)}~Adding target-modal examples improves generalizability but worsens undetectability, causing low ASRD.
  \textbf{(c)}~Our \emph{Proxy Targeted Attack} uses both source-modal and target-modal proxies to keep the AE close to benign data while remaining aligned with cross-modal targets, achieving high ASR and ASRD.
  \vspace{-5mm}
  }
  \label{fig:intro}
\end{figure*}

In addition to limited generalizability, we observe that the Illusion Attack also pushes AE embeddings outside the benign data manifold (see \cref{fig:intro}(a)), making them susceptible to anomaly detection methods \cite{angiulli2002fastknn, breunig2000lof, liu2008isolation, hoffmann2007kernelpca}.
Furthermore, attempts to improve generalizability using multiple target examples widen the discrepancy between AE embeddings and the source-modal reference embeddings, making AEs even more conspicuous (see \cref{fig:intro}(b)).  
Hence, it is challenging to craft a multimodal AE that is both generalizable and undetectable.

In this paper, we aim to improve both the generalizability and undetectability of targeted adversarial examples. 
Specifically, we theoretically explore their underlying connection and propose a novel method, called \textbf{P}roxy \textbf{T}argeted \textbf{A}ttack (PTA). 
PTA leverages not only target-modal proxies but also source-modal proxies to ensure that AEs are sufficiently similar to the latent target, while simultaneously concealing them within source-modal peers.
As a result, PTA improves both the generalizability and undetectability of AEs in cross-modal alignment tasks (see \cref{fig:intro}(c)). 
Comprehensive theoretical analysis and experimental results demonstrate that PTA significantly enhances generalizability and the undetectability of AEs in cross-modal alignment tasks, advancing both the practicality and evasiveness of targeted adversarial attacks on multimodal models.

\section{Related Work}
\noindent\textbf{Multimodal pre-trained models.}
Multimodal pre-trained models \cite{xu2023multimodal, zong2024self, baltruvsaitis2018multimodal, zhu2024vision+} aim to align heterogeneous modalities such as vision, language, and audio into a shared representation space.
Early dual-encoder frameworks, notably CLIP \cite{radford2021learning} and ALIGN \cite{jia2021scaling}, demonstrated strong zero-shot generalization by contrastive training on large-scale noisy image–text pairs, typically using ViT-based visual encoders \cite{dosovitskiy2021image}.
Subsequent works enhanced cross-modal interaction via fusion mechanisms, including ALBEF \cite{li2021align} and BLIP \cite{li2022blip}.
The advent of large vision-language models (LVLMs) further extended this paradigm by connecting frozen vision encoders with large language models, such as Flamingo \cite{alayrac2022flamingo}, BLIP-2 \cite{li2023blip2a}, and LLaVA \cite{liu2023visual}.
Other representative advances include unified training objectives in CoCa \cite{yu2022coca}, scalable contrastive learning in SigLIP \cite{zhai2023sigmoid}, and large-scale vision foundations such as InternVL \cite{chen2024internvl}.
Beyond vision-language alignment, ImageBind \cite{girdhar2023imagebind}, LanguageBind \cite{zhu2024languagebind}, and One-PEACE \cite{wang2023onepeace} further generalized multimodal alignment to multiple sensory modalities.

\noindent\textbf{Adversarial attacks on multimodal models.}
The vulnerability of multimodal models to adversarial attacks has been widely studied \cite{tu2023unicorns,vatsa2023adventures,liu2024safety,fan2024unbridled,zhao2024evaluating,schlarmann2023adversarial,wei2023adaptive,wei2023unified,li2025attack}.
Early attacks focused on untargeted perturbations, including independent modality attacks (e.g., Sep-Attack \cite{madry2019,li-etal-2020-bert-attack}) and joint cross-modal attacks such as Co-Attack \cite{zhang2022adversariala}, SGA \cite{lu2023setlevel}, and CMI-Attack \cite{fu2024improvingadversarialtransferabilityvisionlanguage}.
Targeted attacks were later explored through cosine-similarity optimization \cite{zhang2024adversarial} and surrogate-based objectives \cite{dou2024adversarialattacksmultimodalmodels,zhao2024evaluating}.
With the rise of LVLMs, new jailbreak-style attacks emerged, including universal visual triggers \cite{qi2024visual}, typographic image attacks \cite{gong2024figstep}, and behavior-matching image hijacks \cite{bailey2024image}.
{\color{highlightcolor}
Recent studies further explored compositional attacks \cite{shayegani2024jailbreak,xie2025chain} and large-scale attack settings \cite{zhang2025anyattack}.
}
Notably, Carlini et al. \cite{carlini2023aligned} revealed that safety-aligned language models remain fundamentally vulnerable to adversarial manipulation.
{\color{highlightcolor}
We would like to emphasize that multi-sample, multi-model, and transformation-based strategies are widely used in adversarial learning, mainly to improve cross-model transferability \cite{gu2023survey,dong2018boosting,liu2016delving,xie2019improving,dong2019evading} or robustness to input changes \cite{athalye2018synthesizing,brown2017adversarial,eykholt2018robust,moosavi2017universal}.
Our setting focuses on a different source of uncertainty, which is unique in multimodal systems: the target may be a distribution of possible prompts, captions, retrieval queries, or audio descriptions induced by partial prior knowledge, rather than a fixed label or embedding.
PTA addresses this setting by using target-modal proxies to cover possible cross-modal targets and source-modal proxies to keep the adversarial embedding close to the benign source-modal manifold for anomaly-detection undetectability.
}

\noindent\textbf{Defense mechanisms for multimodal models.}
To mitigate adversarial vulnerabilities, existing defenses mainly rely on adversarial fine-tuning \cite{zhao2024survey,liu2024survey}, including prompt-based adaptation \cite{li2024onea,zhang2024adversariala,mao2023understanding} and full-parameter fine-tuning \cite{wang2024pretraineda,schlarmann2024robusta,waseda2024leveraging,zhou2024revisitinga}.
Representative methods such as TeCoA \cite{mao2023understanding} and FARE \cite{schlarmann2024robusta} improve robustness through contrastive or unsupervised adversarial training.
Other approaches focus on robustness analysis \cite{tu2023closer} and attention alignment \cite{tgazsr2024}.
For LVLM safety, dedicated fine-tuning and randomized smoothing defenses \cite{robey2024smoothllm} have been proposed, alongside standardized benchmarks \cite{liu2024mmsafetybench,tu2023unicorns}.
Nevertheless, existing defenses often suffer from a robustness–accuracy trade-off \cite{mao2023understanding,zhang2019theoretically,raghunathan2019adversarial}.
In contrast, our work emphasizes adversarial example detection as a complementary defense strategy, aiming to reduce harmful influence without degrading benign performance.

\section{Analysis of the Two Limitations} 

This section introduces the threat model and defines generalizability and undetectability of targeted AEs, exposing limitations of previous work.

\subsection{Threat Model}
In this part, we formalize the capabilities and objectives of the adversary in cross-modal matching tasks (i.e., classification and retrieval). 
Let us first denote by \(\mathcal{D}_{\text{S}}\) the data distribution for a source modality and \(\mathcal{D}_{\text{T}}\) the corresponding target modality distribution.

\noindent\textbf{Adversary's objective.}
The adversary's objective is to manipulate the sample $\mathbf{x}$ within an $\epsilon$-ball to generate the targeted adversarial example $\mathbf{x}_{\delta}$ that can mislead the model's matching output toward a desired target \(\mathbf{y}_{t} \sim \mathcal{D}_{\text{T}}\) (i.e., a class prompt in classification and a user query in retrieval). 
Formally, the adversary aims to maximize the \emph{matching score}:
$\text{\large $\tau$}(f_{\theta_\text{S}}(\mathbf{x}_\delta), f_{\theta_\text{T}}(\mathbf{y}_t))$,
where $f_{\theta_\text{S}}$ and $f_{\theta_\text{T}}$ represent the multimodal encoders for the source and target modalities, respectively. 
$\text{\large $\tau$}$ denotes the matching measure (cosine similarity in this paper). 
Moreover, the adversary should consider the effectiveness of AEs under possible defenses such as anomaly detection.

\noindent\textbf{Adversary's capability.}
The adversary can select examples from the source modality and generate AEs from them.
Unlike in traditional classifiers, where the target is a fixed class label, multimodal pre-trained models involve targets from rich modalities (e.g., text or image), which are inherently dynamic and semantically rich.
For example, in classification task, while the target class is static in the closed label space of traditional supervised classifier, multimodal pre-trained models allow downstream practitioners to dynamically specify class prompts (e.g., ``A photo of an elephant'' or ``A huge elephant'') behind one particular class (e.g., elephant).
Thus, the prior assumption that the exact target is accessible to the adversary becomes unrealistic in multimodal systems: in practice, the adversary lacks direct access to the true target \(\mathbf{y}_{t}\) and relies on limited prior knowledge about it. 
{\color{highlightcolor}The concrete source-modal and target-modal proxy construction protocol used to instantiate this limited prior is described in \cref{subsec:proxy_construction}.}
This scenario necessitates attacks generalizable to targets not precisely known to the adversary.
{\color{highlightcolor2}\linelabel{ll:c32:s}For instance, a malicious seller on Amazon OpenSearch knows that prospective buyers will query around the concept of ``handbag'' but cannot predict the precise wording (e.g., ``red leather handbag'' vs.\ ``lightweight canvas tote'').
By crafting and injecting a generalizable adversarial example targeting this broad concept, the seller can surface its products for a wide range of handbag-related queries.\linelabel{ll:c32:e}}

\noindent\textbf{Practical scenarios.}
Our threat model addresses realistic settings where adversaries possess only coarse knowledge of target concepts rather than exact prompt formulations. 
In \emph{content moderation systems} (example of classification), classifiers use private textual prompts (e.g., "a photo of a pistol") to flag sensitive content; an attacker who crafts AEs generalizing across plausible prompt variations can cause a handgun image to be misclassified as benign (e.g., "a person"), thereby bypassing safety filters. 
In \emph{multimodal search systems} (e.g., Amazon OpenSearch with Titan Embeddings) (example of retrieval), sellers upload product images to open catalogs where users issue free-form text queries; an adversarial seller can inject AEs optimized toward broad concepts (e.g., handbag), causing legitimate queries to retrieve counterfeit or misleading items. 
In both scenarios, the key challenge is that downstream prompts or user queries are specified dynamically and privately, requiring AEs to generalize beyond any single known formulation, precisely the capability that this paper investigates.

\subsection{Insufficient Generalizability of the Targeted Attack}   

\label{subsec:generalizability_paradigm}

In this part, we define the concept of generalizability for targeted AEs in cross-modal matching tasks and analyze the limitations of the existing method in this aspect.

Although the adversary does not have direct access to the true target $\mathbf{y}_{t}$, we assume there exists a prior knowledge $Q$ which can be used to define a potential distribution of true targets, denoted as $\mathcal{P}_{\text{target}}(\mathbf{Y} \sim \mathcal{D}_{\text{T}} \vert Q)$, such that $\mathbf{y}_{t} \sim \mathcal{P}_{\text{target}}(\mathbf{Y} \sim \mathcal{D}_{\text{T}} \vert Q)$.
The goal for $\mathbf{x}_{\delta}$ is 
\begin{figure}
  \centering
  \includegraphics[scale=.4]{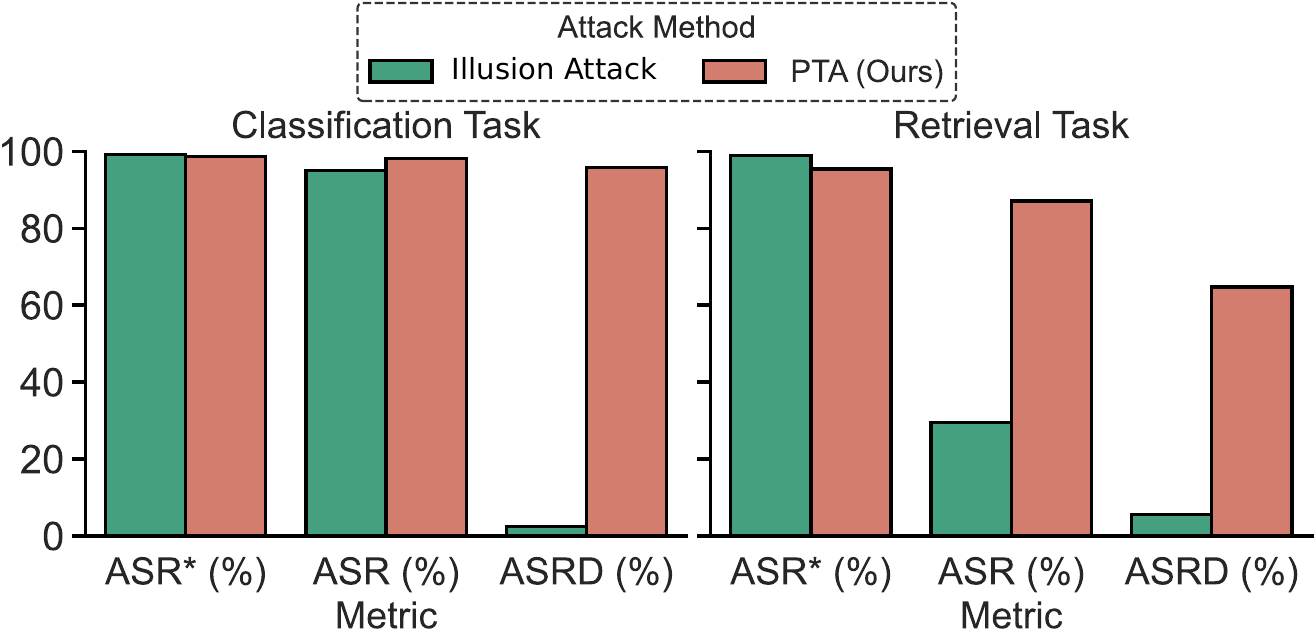}
  \caption{
  Comparison of attack performance in terms of \textbf{generalizability and undetectability}. 
  ASR* represents the attack success rate when the true target is known (\textbf{ideal situation}), ASR corresponds to the attack success rate when the true target is unknown (measures \textbf{generalizability}), and {\color{highlightcolor}ASRD denotes the fraction of all attack attempts that both bypass anomaly detection and succeed when the true target is unknown (further measures \textbf{undetectability}).}
  }
  \vspace{-5mm}
  \label{fig:abilities_compare}
\end{figure}
to generalize across possible samples within $\mathcal{P}_{\text{target}}(\mathbf{Y} \sim \mathcal{D}_{\text{T}} \vert Q)$. 
Thus, the generalizability of AEs can be measured as:
\begin{align}
\nonumber
\mathbb{E}_{\mathbf{y} \sim \mathcal{P}_{\text{target}}(\mathbf{Y} \sim \mathcal{D}_{\text{T}} \vert Q)}\left[\text{\large $\tau$}(f_{\theta_\text{S}}(\mathbf{x}_\delta), f_{\theta_\text{T}}(\mathbf{y}))\right].
\end{align}
For example, if the adversary lacks knowledge of the precise caption in a text-to-image retrieval task, the adversarial image should be able to deceive semantically similar textual descriptions that match certain known keywords. 
For clarity, \emph{generalizability} here refers to the ability of AEs to match partially known or semantically similar cross-modal targets. 
This differs from \emph{transferability}, which measures the ability of AEs generated for one model to also fool another \cite{gu2023survey}.

In \cref{fig:abilities_compare}, we illustrate the ASR of existing targeted attacks to match multiple semantically similar cross-modal targets. 
The results show that AEs crafted by the current method struggle to effectively align with multiple similar targets, which restricts their applicability in practical scenarios. 
This observation motivates us to explore strategies for improving the generalizability of AEs.

\subsection{Limited Undetectability of the Targeted Attack}  

\label{subsec:Undetectability_and_AD_paradigm}
In what follows, we provide a general framework for detecting AEs in the embedding space and analyze adversarial undetectability using anomaly detection.

To well quantify the undetectability of AEs, we summarize a detection framework to identify the outliers likely to be adversarial. The detected outliers can be formalized as:
\begin{align}
\nonumber
\mathbf{D}_{\text{outlier}} = \left\{\mathbf{x}_i \mid s_i > \text{Quantile}(S, 1-r)\right\}_{i=1}^{N},  
\end{align}
where $r$ is a pre-given anomaly ratio in unsupervised anomaly detection, $S = \{s_1, s_2, \dots, s_N\}$ denotes a set of anomaly scores, and $\mathrm{Quantile}(\cdot, 1 - r)$ returns the value at the specified quantile of the anomaly scores. 
It is noteworthy that the calculation of the anomaly score $s$ varies depending on the detection method employed \cite{angiulli2002fastknn, breunig2000lof, hoffmann2007kernelpca, liu2008isolation}, identifying those with higher scores as outliers. 
\begin{table}
\centering
\caption{
\textbf{Average anomaly score ranks} ($\uparrow\%$) of AEs predicted by anomaly detectors (with 100 reference points used). 
The AEs were generated for LanguageBind \cite{zhu2024languagebind}. 
\textbf{A larger rank indicates higher undetectability.}
}
\label{tab:detection_results}
\renewcommand{\arraystretch}{1.1}
\resizebox{0.49\textwidth}{!}{
\setlength{\tabcolsep}{1mm}{
\begin{tabular}{cccccc}

\toprule[1.1pt]

\textbf{Task}                & \textbf{Attack} & \textbf{$k$NN}   & \textbf{LOF}   & \textbf{Forest} & \textbf{PCA}   \\ 
\cmidrule(lr){1-2}\cmidrule(lr){3-6}
\multirow{2}{*}{Classification} & Illusion Attack  & 1.00           & 1.01           & 1.00            & 1.00           \\
                    & \cellcolor{Gray}\textbf{PTA (Ours)}         & \cellcolor{Gray}\textbf{35.64} & \cellcolor{Gray}\textbf{21.78} & \cellcolor{Gray}\textbf{71.29}  & \cellcolor{Gray}\textbf{28.71} \\  \hline
\multirow{2}{*}{Retrieval} & Illusion Attack  & 1.01           & 1.03           & 1.10            & 1.07           \\
                    & \cellcolor{Gray}\textbf{PTA (Ours)}         & \cellcolor{Gray}\textbf{9.30}  & \cellcolor{Gray}\textbf{8.71}  & \cellcolor{Gray}\textbf{3.07}   & \cellcolor{Gray}\textbf{6.18}  \\
\bottomrule[1.1pt]
            
\end{tabular}
}
}
  \vspace{-5mm}
\end{table}

Illusion Attack generates embeddings that lie far outside the benign data manifold (see \cref{fig:intro}(a)), making them vulnerable to anomaly detectors.
In \cref{tab:detection_results}, we show that existing targeted AEs can be effectively detected using simple anomaly detection methods such as $k$NN \cite{angiulli2002fastknn}, LOF \cite{breunig2000lof}, Isolation Forest \cite{liu2008isolation}, and PCA \cite{hoffmann2007kernelpca} in the embedding space, leading to a low Attack Success Rate (ASR) after anomaly detection (ASRD) as illustrated in \cref{fig:abilities_compare}.
This observation motivates us to explore methods to enhance the undetectability of AEs.

\subsection{The Relationship between the Generalizability and Undetectability}
\label{sec:Trade-off_Analysis_theorem}
Here, we analyze how adversaries can generate AEs that achieve both high undetectability and generalizability in multimodal models. 
Since AEs lose their effectiveness once detected as anomalies, the adversary aims to maximize generalizability while remaining as undetectable as possible. 
If adversaries model the defender's anomaly detection algorithm in the embedding space as a distance-based outlier filtering problem \cite{angiulli2002fastknn}, the adversary's optimization objective for AE generation under anomaly detection defense can be formally defined as:
\begin{align}
  \nonumber
  &\min_{\mathbf{x}_\delta} \mathbb{E}_{\mathbf{y} \sim \mathcal{P}_{\text{target}}(\mathbf{Y} \sim \mathcal{D}_{\text{T}} \vert Q)} \left[ d \left(f_{\theta_\text{S}}(\mathbf{x}_{\delta}), f_{\theta_\text{T}}(\mathbf{y}) \right) \right] \\ \nonumber
  &\ \text{s.t.} \ \mathbb{E}_{\mathbf{x} \sim \mathcal{P}_{\text{target}}(\mathbf{X} \sim \mathcal{D}_{\text{S}} \vert Q)} \left[ d \left(f_{\theta_\text{S}}(\mathbf{x}_{\delta}), f_{\theta_\text{S}}(\mathbf{x}) \right) \right] \leq \beta_{\text{true}},
\end{align}
where $d(\cdot)$ denotes a distance metric used in the anomaly detection algorithm, and $\beta_{\text{true}}$ is a threshold that distinguishes benign from anomalous examples. 
{\color{highlightcolor}The objective aims to maximize the generalizability of $\mathbf{x}_{\delta}$ under an anomaly-detection constraint.} However, in practice, adversaries do not know the exact value of $\beta_{\text{true}}$ set by the defender. 
Instead, they can only estimate $\beta$ as an approximation of the detection threshold. 
{\color{highlightcolor}For analytical convenience, we employ the squared-$L_2$ distance as a source-proximity surrogate and analyze the following embedding-space relaxation, where $\mathbf{v}$ corresponds to $f_{\theta_\text{S}}(\mathbf{x}_{\delta})$ but is optimized as a surrogate variable:}
{\color{highlightcolor}
\begin{equation}
\label{eq:opt_obj}
\begin{aligned}
  &\min_{\mathbf{v}} \mathbb{E}_{\mathbf{y}} \left[\|\mathbf{v} - f_{\theta_\text{T}}(\mathbf{y}) \|_2^2\right] \\
  &\ \text{s.t.} \ \mathbb{E}_{\mathbf{x}} \left[\|\mathbf{v} - f_{\theta_\text{S}}(\mathbf{x}) \|_2^2\right] \leq \beta.
\end{aligned}
\end{equation}
}
To solve \cref{eq:opt_obj}, we first introduce a useful lemma.

\begin{Lemma}
\label{lem:bias_variance}
For a random vector $\mathbf{y}$ and a given vector $\mathbf{x}$:
$\mathbb{E}_{\mathbf{y}} [ \|\mathbf{x} - \mathbf{y} \|_2^2 ] = \|\mathbf{x} - \mathbb{E}_{\mathbf{y}}[\mathbf{y}] \|_2^2 + \mathrm{tr}(\mathrm{Var} [\mathbf{y}])$.
\end{Lemma}

This follows by adding and subtracting $\mathbb{E}_{\mathbf{y}}[\mathbf{y}]$, expanding the square, and noting that the cross term vanishes since $\mathbb{E}_{\mathbf{y}}[\mathbb{E}_{\mathbf{y}}[\mathbf{y}] - \mathbf{y}] = 0$.
Based on Lemma~\ref{lem:bias_variance}, we establish the following theorem:

\begin{theo}
\label{theo_1}
{\color{highlightcolor}Let $\mathbf{v}$ denote the relaxed embedding-space variable corresponding to $f_{\theta_\text{S}}(\mathbf{x}_{\delta})$, and define generalizability $L(\mathbf{v}) = \mathbb{E}_{\mathbf{y}} [\|\mathbf{v}- f_{\theta_\text{T}}(\mathbf{y}) \|_2^2]$.}
{\color{highlightcolor}Assume $\beta\ge\sigma_{\text{S}}$ so that the source-proximity constraint is feasible.}
{\color{highlightcolor}We have:}
\begin{gather}
  \nonumber
  \min_{\mathbf{v}} L(\mathbf{v}) = \left( \max \left\{\| \mathbf{\Delta} \|_2 - \sqrt{\beta-\sigma_\text{S}}, 0\right\}\right)^2 + \sigma_\text{T},
\end{gather}
where $\sigma_{\text{T}}=\mathrm{tr}(\mathrm{Var}[f_{\theta_\text{T}}(\mathbf{y})])$, $\sigma_{\text{S}}=\mathrm{tr}(\mathrm{Var}[f_{\theta_\text{S}}(\mathbf{x})])$, and $\mathbf{\Delta} = \mathbb{E}_{\mathbf{y}}[f_{\theta_\text{T}}(\mathbf{y})] - \mathbb{E}_{\mathbf{x}}[f_{\theta_\text{S}}(\mathbf{x})]$ represents the modality gap~\cite{liang2024mind}.  
\end{theo}

\begin{proof}
For notational convenience, let $\mathbb{E}_{\mathbf{x}} \triangleq \mathbb{E}_{\mathbf{x} \sim \mathcal{P}_{\text{target}}(\mathbf{X} | Q)}$ and $\mathbb{E}_{\mathbf{y}} \triangleq \mathbb{E}_{\mathbf{y} \sim \mathcal{P}_{\text{target}}(\mathbf{Y} | Q)}$. 
{\color{highlightcolor}Applying Lemma~\ref{lem:bias_variance} to both the objective and constraint in \cref{eq:opt_obj}, and letting $\boldsymbol{\mu}_{\text{T}}=\mathbb{E}_{\mathbf{y}}[f_{\theta_\text{T}}(\mathbf{y})]$ and $\boldsymbol{\mu}_{\text{S}}=\mathbb{E}_{\mathbf{x}}[f_{\theta_\text{S}}(\mathbf{x})]$, we obtain:}
\begin{align}
  \nonumber
  \min_{\mathbf{v}} \|\mathbf{v} - \boldsymbol{\mu}_{\text{T}} \|_2^2 + \sigma_\text{T}, \quad
  \text{s.t.} \   \|\mathbf{v} - \boldsymbol{\mu}_{\text{S}} \|_2^2 + \sigma_\text{S} \leq \beta.
\end{align}
Using the Lagrangian method with multiplier $\lambda \geq 0$ and slack variable $m$:
\begin{align}
\nonumber
\mathcal{L}(\mathbf{v}, \lambda, m) & = \|\mathbf{v} - \boldsymbol{\mu}_{\text{T}} \|_2^2 + \sigma_\text{T} \\
\nonumber
& \quad + \lambda (\|\mathbf{v} - \boldsymbol{\mu}_{\text{S}} \|_2^2 + \sigma_\text{S} + m^2 -\beta).
\end{align}
Setting $\nabla_{\mathbf{v}} \mathcal{L} = 0$ and analyzing the KKT conditions yields:
\begin{align}
  \nonumber
\mathbf{v}^\star = \begin{cases}
    \boldsymbol{\mu}_\text{S} + \sqrt{\beta-\sigma_\text{S}} \cdot \frac{\mathbf{\Delta}}{\| \mathbf{\Delta} \|_2} & \| \mathbf{\Delta} \|_2 > \sqrt{\beta-\sigma_\text{S}}, \\
   \boldsymbol{\mu}_\text{T} & \| \mathbf{\Delta} \|_2 \le \sqrt{\beta-\sigma_\text{S}},
  \end{cases}
\end{align}
where $\mathbf{\Delta} = \boldsymbol{\mu}_\text{T} - \boldsymbol{\mu}_\text{S}$ denotes the modality gap. Substituting $\mathbf{v}^\star$ into the objective completes the proof.
\end{proof}

{\color{highlightcolor}Theorem~\ref{theo_1} shows that the achievable generalizability $L(\mathbf{v})$ depends on the modality gap $\left\| \mathbf{\Delta} \right\|_2$ and the estimated source-proximity threshold $\beta$.}
{\color{highlightcolor}The actual perturbation-constrained AE optimization in input space may only approximate this relaxed embedding-space solution.}
{\color{highlightcolor}A smaller modality gap or a looser source-proximity threshold reduces $L(\mathbf{v})$, suggesting a surrogate-level trade-off between target alignment and source proximity.}
{\color{highlightcolor}Thus, optimizing only toward target-modal data can improve target alignment but may move the AE away from the benign source-modal region, as illustrated in \cref{fig:intro}(b).}
{\color{highlightcolor}Therefore, incorporating source-modal proxies provides a practical way to regularize this movement and improve undetectability under anomaly-detection defenses.}

{\color{highlightcolor2}
\noindent\textbf{Scope of the $L_2$ source-proximity analysis.}
Theorem~\ref{theo_1} should be interpreted as an analysis of the $L_2$ source-proximity surrogate used by PTA, rather than as a detector-exact guarantee for every anomaly detector.
The theorem uses
\[
D_{\mathrm{S}}(\mathbf{v})
=
\mathbb{E}_{\mathbf{x}}
\left[
\|\mathbf{v}-f_{\theta_\text{S}}(\mathbf{x})\|_2^2
\right]
\le \beta
\]
as a tractable notion of undetectability.
Intuitively, a small $D_{\mathrm{S}}(\mathbf{v})$ means that the AE embedding $\mathbf{v}$ is not an isolated point in the source-modal embedding space, but stays near benign source-modal samples.
By Markov's inequality, this condition also implies a local-mass bound:
\[
\Pr_{\mathbf{x}}\left(\|\mathbf{v}-f_{\theta_\text{S}}(\mathbf{x})\|_2\le r\right)
\ge
1-\frac{\beta}{r^2},\quad r^2>\beta .
\]
Thus, although Theorem~\ref{theo_1} does not model the exact score of every detector, it explains why keeping $D_{\mathrm{S}}(\mathbf{v})$ small is relevant to detectors that search for isolated or low-density embeddings.

For $k$NN, the connection is direct.
Let $\widehat{D}_{\mathrm{S}}(\mathbf{v})=\frac{1}{n}\sum_{i=1}^{n}\|\mathbf{v}-\mathbf{z}_i\|_2^2$ be the empirical source-proximity statistic and let $d_{(k)}(\mathbf{v})$ be the $k$-th nearest-neighbor distance to benign references.
Then
\[
d_{(k)}^2(\mathbf{v})
\le
\frac{n}{n-k+1}\widehat{D}_{\mathrm{S}}(\mathbf{v}),
\]
so reducing $\widehat{D}_{\mathrm{S}}(\mathbf{v})$ also controls a sufficient upper bound on the $k$NN anomaly score.
For PCA, if the anomaly score is the reconstruction residual,
\[
s_{\mathrm{PCA}}(\mathbf{v})
=
\|(I-UU^\top)(\mathbf{v}-\hat{\boldsymbol{\mu}}_{\mathrm{S}})\|_2^2
\le
\|\mathbf{v}-\hat{\boldsymbol{\mu}}_{\mathrm{S}}\|_2^2,
\]
so an embedding close to the benign source distribution also has a conservative upper bound on its PCA residual.
For LOF, the connection is indirect but still meaningful: LOF is computed from neighbor distances and local reachability density, so source proximity increases the local benign density around the AE under standard local-density regularity assumptions.
For Isolation Forest, which relies on random recursive partitioning rather than explicit distances, Theorem~\ref{theo_1} does not imply a deterministic guarantee.
It only provides a conditional intuition: an AE surrounded by benign source-modal neighbors is less likely to be isolated by very shallow random splits.

The empirical anomaly-rank results for $k$NN, LOF, Isolation Forest, and PCA are reported in \cref{tab:detection_results}.
Accordingly, Theorem~\ref{theo_1} motivates the $L_2$ source-proximity objective, with direct or conditional connections to the detectors discussed above.
For detectors that are not strictly covered by the theorem, these anomaly-rank results provide empirical justification for applying the source-proximity objective to practical anomaly detectors such as $k$NN, LOF, Isolation Forest, and PCA.
}

{\color{highlightcolor2}
\linelabel{ll:c31a:s}Theorem~\ref{theo_1} involves two approximation gaps.
First, it optimizes a free embedding vector $\mathbf{v}$, while PTA operates under input-space $\ell_\infty$ constraints.
PTA nonetheless achieves an average of 98.31\% Cls ASR and 77.72\% R@1 ASR (\cref{tab:classification_whitebox,tab:retrieval_whitebox}), indicating that input-space optimization can reach near-optimal embedding-space solutions despite this constraint, and the $\alpha$ ablation in \cref{fig:ABL-alpha-ASR_D-and-ASR_retri} reproduces the generalizability/undetectability trade-off predicted by the theorem.
Second, the theorem uses squared-$L_2$ source proximity as the undetectability surrogate, while defenders deploy diverse anomaly detectors.
\cref{tab:detection_results} shows that optimizing this $L_2$ surrogate consistently increases the anomaly-score rank percentiles toward the benign region across all four detector families, confirming that the surrogate transfers to practical detection metrics.\linelabel{ll:c31a:e}
}

{\color{highlightcolor}
Based on this observation, we design a new attack method that jointly considers target generalizability and source-modal undetectability of AEs in the next section.
}

\section{PTA: Proxy Targeted Attack}
\label{sec:New_Attack_Method}

\subsection{\color{highlightcolor}Method overview}
\label{subsec:pta_method_overview}

In \cref{subsec:Undetectability_and_AD_paradigm,subsec:generalizability_paradigm}, we identified the limitations of the existing targeted attack on undetectability and generalizability. Furthermore, \cref{sec:Trade-off_Analysis_theorem} presents a theoretical analysis that reveals a limitation: these two challenges cannot be perfectly addressed simultaneously.
To address these challenges, we propose Proxy Targeted Attack (PTA), endowing the AEs with both generalizability and undetectability. 
It introduces two key innovations: (i) leveraging multiple proxy targets to enhance generalizability, and (ii) optimizing AEs with respect to source-modal targets to improve undetectability.

To improve generalizability, we define the optimization loss $\mathcal{L}_{\text{G}}$ with $\mathbf{x}_{\delta}$ as:
\begin{equation}
\label{eq:generalization_loss}
\mathcal{L}_{\text{G}}(\mathbf{x}_{\delta}) = 1 - \frac{1}{N_{c}} \sum\nolimits_{j=1}^{N_{c}} \text{\large $\tau$} \left(f_{\theta_{\text{S}}}(\mathbf{x}_{\delta}), f_{\theta_{\text{T}}}(\hat{\mathbf{y}}_{j})\right),
\end{equation}
where $\{\hat{\mathbf{y}}_{1}, ..., \hat{\mathbf{y}}_{N_c}\}$ denotes a set of \textit{target-modal proxies} that are sampled from the estimated ground-truth distribution $\mathcal{P}_{\text{target}}(\mathbf{Y} \sim \mathcal{D}_{\text{T}} \vert Q)$. Note that $N_c$ is a hyperparameter.
We utilize this set of proxy targets to serve as surrogates for the unknown true target $\mathbf{y}_{t}$. By maximizing the cosine similarity, we can enhance the generalizability of AEs across multiple cross-modal targets, thereby increasing the likelihood of successful attacks on the true targets.

To improve undetectability, we define the undetectability optimization loss $\mathcal{L}_{\text{D}}$ for $\mathbf{x}_{\delta}$ as:
\begin{equation}
\label{eq:undetectability_loss}
\mathcal{L}_{\text{D}}(\mathbf{x}_{\delta}) = \frac{1}{N_{s}} \sum\nolimits_{i=1}^{N_{s}} \left\|f_{\theta_{\text{S}}}(\mathbf{x}_{\delta}) - f_{\theta_{\text{S}}}(\hat{\mathbf{x}}_{i})\right\|_2,
\end{equation}
where $\{\hat{\mathbf{x}}_{1}, ..., \hat{\mathbf{x}}_{N_s}\}$ denotes a set of \textit{source-modal proxies} that is sampled from the estimated distribution $\mathcal{P}_{\text{target}}(\mathbf{X} \sim \mathcal{D}_{\text{S}} \vert Q)$ and $N_{s}$ is the number of source-modal proxies.
The objective is to position the AEs as close as possible to these source-modal proxies in the embedding space. Ideally, the AEs should lie within or near the convex polytope formed by benign examples, thereby enhancing concealment and minimizing the likelihood of detection. 

Combining \cref{eq:generalization_loss} and \cref{eq:undetectability_loss}, the final optimization objective is defined as:
\begin{equation}
    \mathop{\arg \min}\limits_{\mathbf{x}_\delta} \mathcal{L}_{\text{G}}(\mathbf{x}_{\delta}) + \alpha \mathcal{L}_{\text{D}}(\mathbf{x}_{\delta}), \,\, \text{s.t.} ,\, \Vert \mathbf{x}_\delta\ - \mathbf{x} \Vert_{\infty} \leq \epsilon,
\label{eq:overall_loss}
\end{equation}
where $\mathbf{x}$ denotes the original example corresponding to $\mathbf{x}_{\delta}$ and the perturbation is constrained by a maximum perturbation limit $\epsilon$. The parameter $\alpha$ serves as a balancing factor between $\mathcal{L}_{\text{G}}$, and $\mathcal{L}_{\text{D}}$, allowing control over the dominance of the two abilities. 

\subsection{\color{highlightcolor}Theoretical analysis of the PTA losses}
\label{subsec:pta_loss_theory}

{\color{highlightcolor}
\noindent\textbf{Geometric effect of source-modal proxies.}
The source-modal loss in \cref{eq:undetectability_loss} pulls the AE embedding toward benign source-modal proxies and acts as a source-manifold regularizer.
We first provide a sufficient lower-bound view of this regularization: if the AE embedding stays inside or near the convex hull of the source-modal proxies, its similarity to the true target admits a lower bound.

\begin{theo}
\label{thm:theo_2}
Let
\[
\mathbf{v}=f_{\theta_{\mathrm{S}}}(\mathbf{x}_{\delta}),\quad
\hat{\mathbf{v}}_i=f_{\theta_{\mathrm{S}}}(\hat{\mathbf{x}}_i),\quad
\mathbf{t}=f_{\theta_{\mathrm{T}}}(\mathbf{y}_t),
\]
where all embeddings are $\ell_2$-normalized for cosine similarity.
Let
\[
\mathcal{P}_{\mathrm{S}}=\mathrm{conv}\{\hat{\mathbf{v}}_i\}_{i=1}^{N_s},
\qquad
h_{\mathrm{S}}(\mathbf{v})=\mathrm{dist}(\mathbf{v},\mathcal{P}_{\mathrm{S}}),
\]
and
\[
B_{N_s}=\min_{i\in[N_s]}\tau(\hat{\mathbf{v}}_i,\mathbf{t}).
\]
Then the following source-proxy lower bound holds:
\begin{align}
\nonumber
\tau(\mathbf{v},\mathbf{t})
\ge
\begin{cases}
B_{N_s}, & h_{\mathrm{S}}(\mathbf{v})=0 \;(\mathbf{v}\in\mathcal{P}_{\mathrm{S}}),\\
B_{N_s}-h_{\mathrm{S}}(\mathbf{v}), & h_{\mathrm{S}}(\mathbf{v})>0 \;(\mathbf{v}\notin\mathcal{P}_{\mathrm{S}}).
\end{cases}
\end{align}
{\color{highlightcolor2}Thus, when $\mathbf{v}$ moves outside the source-proxy hull, the lower bound degrades linearly with the hull distance $h_{\mathrm{S}}(\mathbf{v})$.}
\end{theo}

\begin{proof}
Let $\mathbf{p}^{\star}=\arg\min_{\mathbf{p}\in\mathcal{P}_{\mathrm{S}}}\|\mathbf{v}-\mathbf{p}\|_2$.
Since $\mathbf{p}^{\star}\in\mathcal{P}_{\mathrm{S}}$, there exist coefficients $\beta_i\ge0$ and $\sum_i\beta_i=1$ such that $\mathbf{p}^{\star}=\sum_i\beta_i\hat{\mathbf{v}}_i$.
Therefore,
\[
(\mathbf{p}^{\star})^\top\mathbf{t}
=\sum_i\beta_i\hat{\mathbf{v}}_i^\top\mathbf{t}
\ge B_{N_s}.
\]
Using $\|\mathbf{t}\|_2=1$, we obtain
\begin{align}
\nonumber
\tau(\mathbf{v},\mathbf{t})
&=\mathbf{v}^{\top}\mathbf{t} \\
\nonumber
&=(\mathbf{p}^{\star})^{\top}\mathbf{t}
+(\mathbf{v}-\mathbf{p}^{\star})^{\top}\mathbf{t} \\
\nonumber
&\ge
B_{N_s}-\|\mathbf{v}-\mathbf{p}^{\star}\|_2
=B_{N_s}-h_{\mathrm{S}}(\mathbf{v}).
\end{align}
Since $\mathcal{P}_{\mathrm{S}}$ is closed, $h_{\mathrm{S}}(\mathbf{v})=0$ if and only if $\mathbf{v}\in\mathcal{P}_{\mathrm{S}}$.
The two cases in the theorem follow immediately.
\end{proof}

This result gives a geometric interpretation of $\mathcal{L}_{\mathrm{D}}$: keeping $\mathbf{v}$ close to the source-proxy hull preserves a lower bound on target similarity, and the bound degrades linearly as $\mathbf{v}$ moves away from the hull.
Although \cref{eq:undetectability_loss} does not impose $\mathbf{v}\in\mathcal{P}_{\mathrm{S}}$ as a hard constraint, it serves as a soft hull-proximity regularizer.
Consider its embedding-space form
\[
\widetilde{\mathcal{L}}_{\mathrm{D}}(\mathbf{u})
=
\frac{1}{N_s}\sum_{i=1}^{N_s}\|\mathbf{u}-\hat{\mathbf{v}}_i\|_2.
\]
For any $\mathbf{v}$, let $\mathbf{q}^{\star}=\Pi_{\mathcal{P}_{\mathrm{S}}}(\mathbf{v})$ be its projection onto the source-proxy hull.
Since $\mathcal{P}_{\mathrm{S}}$ is closed and convex, the projection theorem gives
\[
\|\mathbf{q}^{\star}-\hat{\mathbf{v}}_i\|_2
\le
\|\mathbf{v}-\hat{\mathbf{v}}_i\|_2
\quad \text{for every } \hat{\mathbf{v}}_i\in\mathcal{P}_{\mathrm{S}}.
\]
Thus, $\widetilde{\mathcal{L}}_{\mathrm{D}}(\mathbf{q}^{\star})\le \widetilde{\mathcal{L}}_{\mathrm{D}}(\mathbf{v})$, so the source-modal loss favors embeddings close to the source-proxy hull.
In the full PTA objective, $\mathcal{L}_{\mathrm{G}}$ provides target alignment, while $\mathcal{L}_{\mathrm{D}}$ preserves source-manifold proximity without unboundedly sacrificing target similarity.
}

{\color{highlightcolor}
\noindent\textbf{Coverage effect of target-modal proxies.}
The target-modal loss in \cref{eq:generalization_loss} aligns the AE embedding with target-modal proxies, which represent possible targets induced by $Q$.
The following result shows when this proxy alignment transfers to the unknown true target: if the true target embedding lies inside or near the convex hull of the target-modal proxies, its similarity to the AE has a lower bound.

\begin{theo}
  \label{theo_3}
Let
\[
\mathbf{v}=f_{\theta_{\mathrm{S}}}(\mathbf{x}_{\delta}),\quad
\hat{\mathbf{z}}_j=f_{\theta_{\mathrm{T}}}(\hat{\mathbf{y}}_j),\quad
\mathbf{t}=f_{\theta_{\mathrm{T}}}(\mathbf{y}_t),
\]
where all embeddings are $\ell_2$-normalized for cosine similarity.
Let
\[
\mathcal{P}_{\mathrm{T}}=\mathrm{conv}\{\hat{\mathbf{z}}_j\}_{j=1}^{N_c},
\qquad
h_{\mathrm{T}}(\mathbf{t})=\mathrm{dist}(\mathbf{t},\mathcal{P}_{\mathrm{T}}),
\]
and
\[
B_{N_c}=\min_{j\in[N_c]}\tau(\mathbf{v},\hat{\mathbf{z}}_j).
\]
Then the following target-proxy lower bound holds:
\begin{align}
\nonumber
\tau(\mathbf{v},\mathbf{t})
\ge
\begin{cases}
B_{N_c}, & h_{\mathrm{T}}(\mathbf{t})=0 \;(\mathbf{t}\in\mathcal{P}_{\mathrm{T}}),\\
B_{N_c}-h_{\mathrm{T}}(\mathbf{t}), & h_{\mathrm{T}}(\mathbf{t})>0 \;(\mathbf{t}\notin\mathcal{P}_{\mathrm{T}}).
\end{cases}
\end{align}
Thus, when the true target embedding lies outside the target-proxy hull, the lower bound degrades linearly with the target-hull distance $h_{\mathrm{T}}(\mathbf{t})$.
\end{theo}

\begin{proof}
Let $\mathbf{p}^{\star}=\arg\min_{\mathbf{p}\in\mathcal{P}_{\mathrm{T}}}\|\mathbf{t}-\mathbf{p}\|_2$.
Since $\mathbf{p}^{\star}\in\mathcal{P}_{\mathrm{T}}$, there exist coefficients $\gamma_j\ge0$ and $\sum_j\gamma_j=1$ such that $\mathbf{p}^{\star}=\sum_j\gamma_j\hat{\mathbf{z}}_j$.
Therefore,
\[
\mathbf{v}^{\top}\mathbf{p}^{\star}
=\sum_j\gamma_j\mathbf{v}^{\top}\hat{\mathbf{z}}_j
\ge B_{N_c}.
\]
Using $\|\mathbf{v}\|_2=1$, we obtain
\begin{align}
\nonumber
\tau(\mathbf{v},\mathbf{t})
&=\mathbf{v}^{\top}\mathbf{t} \\
\nonumber
&=\mathbf{v}^{\top}\mathbf{p}^{\star}
+\mathbf{v}^{\top}(\mathbf{t}-\mathbf{p}^{\star}) \\
\nonumber
&\ge
B_{N_c}-\|\mathbf{t}-\mathbf{p}^{\star}\|_2
=B_{N_c}-h_{\mathrm{T}}(\mathbf{t}).
\end{align}
Since $\mathcal{P}_{\mathrm{T}}$ is closed, $h_{\mathrm{T}}(\mathbf{t})=0$ if and only if $\mathbf{t}\in\mathcal{P}_{\mathrm{T}}$.
The two cases in the theorem follow immediately.
\end{proof}

Theorem~\ref{theo_3} shows the role of target-modal proxy coverage.
If the target-modal proxies cover the true target embedding, i.e., $h_{\mathrm{T}}(\mathbf{t})=0$, then aligning the AE with all proxies gives a lower bound on its similarity to the true target.
If the true target is outside but close to the target-proxy hull, the lower bound degrades linearly with $h_{\mathrm{T}}(\mathbf{t})$.
Therefore, diverse and representative target-modal proxies reduce $h_{\mathrm{T}}(\mathbf{t})$ and improve generalizability.
Because \cref{eq:generalization_loss} optimizes average proxy similarity rather than $B_{N_c}$ directly, the bound is most informative when the selected proxies are semantically coherent and the optimized AE maintains high similarity to the proxy set.

Together, Theorems~\ref{thm:theo_2} and~\ref{theo_3} explain the different roles of the two proxy types: source-modal proxies keep the AE embedding close to benign source samples, while target-modal proxies help cover possible cross-modal targets.
}

\subsection{\color{highlightcolor2}Why single-target alignment can fail in practice}
\label{subsec:why_illusion_fails}

{\color{highlightcolor}
The proxy-based analysis also clarifies the limitation of single-target alignment.
Illusion Attack optimizes an AE toward one specific target embedding, whereas practical cross-modal tasks involve a target distribution induced by the known concept, category, or partial query $Q$.
Let $\mathbf{z}=f_{\theta_\text{T}}(\mathbf{y})$ be a unit-normalized embedding sampled from $\mathcal{P}_{\text{target}}(\mathbf{Y}\sim\mathcal{D}_{\mathrm{T}}\mid Q)$.
Let $\mathbf{z}_0=f_{\theta_\text{T}}(\mathbf{y}_0)$ be the single target used by Illusion Attack, and let $\mathbf{v}=f_{\theta_\text{S}}(\mathbf{x}_{\delta})$ be the unit-normalized adversarial embedding.
The desired distribution-level generalizability is
\[
G(\mathbf{v})=\mathbb{E}_{\mathbf{z}}[\tau(\mathbf{v},\mathbf{z})]
=\mathbf{v}^{\top}\boldsymbol{\mu}_{\mathrm{T}},
\qquad
\boldsymbol{\mu}_{\mathrm{T}}=\mathbb{E}[\mathbf{z}].
\]
If Illusion Attack succeeds on the optimized target, i.e., $\tau(\mathbf{v},\mathbf{z}_0)\ge 1-\eta$, then $\|\mathbf{v}-\mathbf{z}_0\|_2\le \sqrt{2\eta}$.
Therefore,
\[
G(\mathbf{v})
=\mathbf{z}_0^{\top}\boldsymbol{\mu}_{\mathrm{T}}
+(\mathbf{v}-\mathbf{z}_0)^{\top}\boldsymbol{\mu}_{\mathrm{T}}
\le
\mathbf{z}_0^{\top}\boldsymbol{\mu}_{\mathrm{T}}
+\sqrt{2\eta}.
\]
Thus, single-target success transfers to unseen targets only when $\mathbf{z}_0$ is representative of the target distribution.
When possible target embeddings are dispersed, $\mathbf{z}_0$ may be poorly aligned with valid targets, keeping $G(\mathbf{v})$ low.
In addition, if $\mathbf{z}_0$ is also drawn from the target distribution, then
\[
\mathbb{E}_{\mathbf{z}_0}[G(\mathbf{v})]
\le
\|\boldsymbol{\mu}_{\mathrm{T}}\|_2^2+\sqrt{2\eta}
=
1-\mathrm{tr}(\mathrm{Var}[\mathbf{z}])+\sqrt{2\eta},
\]
where the last equality follows from $\mathrm{tr}(\mathrm{Var}[\mathbf{z}])=1-\|\boldsymbol{\mu}_{\mathrm{T}}\|_2^2$ for normalized embeddings.
In the near-perfect case where $\eta\approx0$, the expected generalizability is therefore controlled by $1-\mathrm{tr}(\mathrm{Var}[\mathbf{z}])$.
This means that larger target-distribution variance makes a single optimized target less representative and single-target alignment less reliable.

In contrast, PTA uses target-modal proxies to optimize an empirical estimate of the same distribution-level objective \(G(\mathbf{v})\), instead of committing to one target embedding.
For target-modal proxies $\{\mathbf{z}_j\}_{j=1}^{N_c}$, PTA maximizes
\[
\widehat{G}_{N_c}(\mathbf{v})
=
\frac{1}{N_c}\sum_{j=1}^{N_c}\tau(\mathbf{v},\mathbf{z}_j)
=
\mathbf{v}^{\top}\bar{\mathbf{z}}_{N_c},
\]
where
\[
\bar{\mathbf{z}}_{N_c}=\frac{1}{N_c}\sum_{j=1}^{N_c}\mathbf{z}_j .
\]
{\color{highlightcolor2}Under representative proxy sampling,}
\[
\mathbb{E}[\bar{\mathbf{z}}_{N_c}]=\boldsymbol{\mu}_{\mathrm{T}},
\qquad
\mathbb{E}\|\bar{\mathbf{z}}_{N_c}-\boldsymbol{\mu}_{\mathrm{T}}\|_2^2
=\frac{\mathrm{tr}(\mathrm{Var}[\mathbf{z}])}{N_c}.
\]
Therefore, increasing the number and coverage of representative target-modal proxies makes the optimized direction better approximate the target distribution, reducing single-target overfitting and improving AE's generalizability.

This gives the methodological distinction between PTA and Illusion Attack.
Illusion Attack performs single-target alignment to one exact embedding $f_{\theta_{\mathrm{T}}}(\mathbf{y}_0)$, while PTA performs target-distribution alignment under a coarse concept, keyword, category, or partial query $Q$.
The target-modal loss $\mathcal{L}_{\mathrm{G}}$ estimates the direction of possible cross-modal targets, and the source-modal loss $\mathcal{L}_{\mathrm{D}}$ regularizes the AE toward the benign source-modal manifold.
The balancing factor $\alpha$ in \cref{eq:overall_loss} then controls the generalizability--undetectability trade-off within a single detection-aware optimization objective.
}

{\color{highlightcolor2}
\linelabel{ll:c22:s}The preceding expectation-based result assumes that the single target is randomly sampled.
Here we analyze the more practical case of deliberate target selection.

The ideal optimization direction for distribution-level generalizability is the population mean direction $\mathbf{u}_{\mathrm{T}} = \boldsymbol{\mu}_{\mathrm{T}} / \|\boldsymbol{\mu}_{\mathrm{T}}\|_2$.
However, $\mathbf{u}_{\mathrm{T}}$, being an average over encoder outputs, generally does not correspond to any single realizable target embedding in practice due to the discreteness of practical inputs.

Formally, for a finite candidate pool $\mathcal{C}_N = \{\mathbf{z}_1, \ldots, \mathbf{z}_N\}$ and a data-driven selection $\hat{\mathbf{z}} = \arg\max_{\mathbf{z}_i} \mathbf{z}_i^{\top}\hat{\boldsymbol{\mu}}$, where $\hat{\boldsymbol{\mu}} = \frac{1}{M}\sum_{j=1}^{M}\tilde{\mathbf{z}}_j$ is the empirical mean of $M$ attacker-collected target embeddings $\tilde{\mathbf{z}}_j$, the total gap decomposes as:
\[
\underbrace{
\|\boldsymbol{\mu}_{\mathrm{T}}\|_2 - \hat{\mathbf{z}}^{\top}\boldsymbol{\mu}_{\mathrm{T}}
}_{\text{total single-target gap}}
=
\underbrace{
\|\boldsymbol{\mu}_{\mathrm{T}}\|_2 - \max_i \mathbf{z}_i^{\top}\boldsymbol{\mu}_{\mathrm{T}}
}_{Q_N:\ \text{representation error}}
+
\underbrace{
\max_i \mathbf{z}_i^{\top}\boldsymbol{\mu}_{\mathrm{T}} - \hat{\mathbf{z}}^{\top}\boldsymbol{\mu}_{\mathrm{T}}
}_{R_N:\ \text{selection error}},
\]
where $Q_N = \frac{\|\boldsymbol{\mu}_{\mathrm{T}}\|_2}{2} \min_i \|\mathbf{z}_i - \mathbf{u}_{\mathrm{T}}\|_2^2 \ge 0$ is the mismatch between the best candidate and the mean direction, and $R_N \le 2\|\hat{\boldsymbol{\mu}} - \boldsymbol{\mu}_{\mathrm{T}}\|_2$ is the error from estimating which candidate is best.
Deliberate selection can reduce $R_N$ but not $Q_N$; conversely, $Q_N$ is considerable in practice.
In contrast, PTA optimizes against the empirical proxy mean $\bar{\mathbf{z}}_{N_c}$ directly, which approximates $\boldsymbol{\mu}_{\mathrm{T}}$ with error shrinking as $O(1/\sqrt{N_c})$ without committing to a single candidate.

Empirically, we strengthen Illusion Attack by allowing centroid-based selection from increasingly large candidate pools (\cref{tab:rev-c22-deliberate}).

Although deliberate selection improves centroid alignment, this translates into only modest and non-monotonic R@1 ASR gains.
PTA, using far fewer proxies, substantially outperforms all single-target variants, showing that optimizing toward the target distribution directly is more effective than committing to any single target.\linelabel{ll:c22:e}

\linelabel{ll:c31b:s}The unbiasedness analysis assumes proxies drawn from the true target distribution, whereas practical proxies may follow a biased proxy distribution whose population mean $\widetilde{\boldsymbol{\mu}}$ deviates from $\boldsymbol{\mu}_{\mathrm{T}}$, yielding a systematic bias $\mathbf{b}=\widetilde{\boldsymbol{\mu}}-\boldsymbol{\mu}_{\mathrm{T}}\ne\mathbf{0}$ and an irreducible error floor $\|\mathbf{b}\|^2$.
The proxy-count ablations in \cref{fig:ABL-numbersOfTargets} and Tables~\ref{tab:proxy_quality_classification}--\ref{tab:proxy_quality_retrieval} confirm this bias-variance structure: performance improves with more proxies but plateaus above zero error, consistent with the theoretical prediction.\linelabel{ll:c31b:e}
}

\begin{table*}[!t]
  \centering
  \caption{Comparison of \textbf{generalizability and undetectability} of AEs in classification task. Performance is reported by \textit{Cls ASR (\%)} and \textit{Cls ASRD (\%)} when anomaly-detection defense is used.}
  \label{tab:classification_whitebox}
\resizebox{0.98\textwidth}{!}{
\setlength{\tabcolsep}{1mm}{
  \begin{tabular}{c c ccc ccc c}
    \toprule[1.1pt]
    &  & \multicolumn{3}{c}{\textbf{XmediaNet}} & \multicolumn{3}{c}{\textbf{ImageNet}} &  \multirow{2}{*}{\textbf{Average}} \\ 
    \cmidrule(lr){3-5} \cmidrule(lr){6-8} 
    \textbf{Attack} & \textbf{Defense} & ImageBind & LanguageBind & One-PEACE & ImageBind & LanguageBind & One-PEACE &  \\ 
    \cmidrule(r){1-2} \cmidrule(lr){3-5} \cmidrule(l){6-8} \cmidrule(lr){9-9}

    MF-ii            & \ding{55}       & $30.89_{0.63}$  & $48.00_{4.71}$  & $36.67_{3.31}$  & $17.33_{15.04}$  & $57.78_{10.75}$  & $32.90_{0.61}$ & $37.26$  \\ 
    CrossFire        & \ding{55}       & $31.33_{0.54}$  & $45.56_{6.49}$  & $38.00_{3.93}$  & $12.89_{6.07}$  & $53.33_{3.93}$  & $29.60_{0.49}$ & $35.12$  \\ 
    Illusion Attack  & \ding{55}      & $\textbf{99.58}_{0.59}$  & $\textbf{100.00}_{0.00}$  & $97.50_{0.00}$  & $77.44_{0.21}$  & $95.06_{0.69}$  & $89.50_{0.64}$  & $93.18$  \\ 
    
    \cellcolor{Gray}\textbf{PTA (Ours)} & \cellcolor{Gray}\ding{55} & \cellcolor{Gray}$\textbf{99.58}_{0.59}$  & \cellcolor{Gray}$\textbf{100.00}_{0.00}$  & \cellcolor{Gray}$\textbf{98.75}_{0.59}$  & \cellcolor{Gray}$\textbf{94.22}_{0.47}$  & \cellcolor{Gray}$\textbf{99.72}_{0.20}$  & \cellcolor{Gray}$\textbf{97.61}_{0.78}$  & \cellcolor{Gray}$\textbf{98.31}$ \\ 
    \cmidrule(r){1-9}

    MF-ii            & \ding{51}      & $20.89_{7.78}$  & $43.69_{2.70}$  & $36.67_{3.31}$  & $6.67_{7.62}$  & $35.33_{2.37}$  & $28.70_{0.52}$  & $28.66$  \\ 
    CrossFire        & \ding{51}      & $31.33_{0.54}$  & $42.16_{4.79}$  & $35.97_{2.74}$  & $9.33_{8.56}$  & $41.78_{9.11}$  & $25.10_{0.40}$  & $30.95$  \\ 
    Illusion Attack  & \ding{51}      & $18.33_{0.00}$  & $18.33_{1.18}$  & $66.25_{0.59}$  & $0.42_{0.59}$  & $2.50_{0.00}$  & $14.42_{0.15}$  & $20.04$  \\ 

    \cellcolor{Gray}\textbf{PTA (Ours)} & \cellcolor{Gray}\ding{51}     
    & \cellcolor{Gray}$\textbf{92.92}_{0.59}$  
    & \cellcolor{Gray}$\textbf{95.42}_{0.59}$  & \cellcolor{Gray}$\textbf{87.92}_{0.59}$  & \cellcolor{Gray}$\textbf{77.54}_{1.31}$  & \cellcolor{Gray}$\textbf{95.82}_{0.73}$  & \cellcolor{Gray}$\textbf{55.42}_{1.33}$  & \cellcolor{Gray}$\textbf{84.17}$ \\ 

    \bottomrule[1.1pt]
  \end{tabular}
  }
  }
\end{table*}

\begin{table*}[t]
  \centering
  \caption{Comparison of \textbf{generalizability and undetectability} of AEs in retrieval task. 
  Performance is reported by \textit{R@1 ASR (\%)} and \textit{R@1 ASRD (\%)} when anomaly-detection defense is used.}
  \label{tab:retrieval_whitebox}
\resizebox{0.98\textwidth}{!}{
\setlength{\tabcolsep}{1mm}{
  \begin{tabular}{c c ccc ccc c}
    \toprule[1.1pt]

    & & \multicolumn{3}{c}{\textbf{XmediaNet}} & \multicolumn{3}{c}{\textbf{MSCOCO}} &  \multirow{2}{*}{\textbf{Average}} \\ 
    \cmidrule(lr){3-5} \cmidrule(lr){6-8} 
    \textbf{Attack} & \textbf{Defense} & ImageBind & LanguageBind & One-PEACE & ImageBind & LanguageBind & One-PEACE & \\ 
    \cmidrule(r){1-2} \cmidrule(lr){3-5} \cmidrule(l){6-8} \cmidrule(lr){9-9}

    MF-ii           & \ding{55} & $0.00_{0.00}$  & $0.00_{0.00}$  & $0.00_{0.00}$  & $0.00_{0.00}$  & $0.00_{0.00}$ & $0.00_{0.00}$  & $0.00$  \\ 
    CrossFire       & \ding{55} & $0.00_{0.00}$  & $0.00_{0.00}$ & $0.00_{0.00}$  & $0.00_{0.00}$  & $0.00_{0.00}$ & $0.00_{0.00}$ & $0.00$  \\ 
    Illusion Attack  & \ding{55} & $77.05_{0.47}$  & $85.80_{1.72}$  & $56.76_{1.33}$  & $20.33_{1.30}$  & $29.41_{0.75}$  & $9.59_{0.18}$  & $46.49$ \\ 
    \cellcolor{Gray}\textbf{PTA (Ours)} & \cellcolor{Gray}\ding{55} & \cellcolor{Gray}$\textbf{95.36}_{0.27}$  & \cellcolor{Gray}$\textbf{96.75}_{0.04}$  & \cellcolor{Gray}$\textbf{85.14}_{0.12}$  & \cellcolor{Gray}$\textbf{71.31}_{0.71}$  & \cellcolor{Gray}$\textbf{87.09}_{0.13}$  & \cellcolor{Gray}$\textbf{30.69}_{0.18}$  & \cellcolor{Gray}$\textbf{77.72}$ \\ 
    \cmidrule(r){1-9}
    MF-ii           & \ding{51} & $0.00_{0.00}$  & $0.00_{0.00}$  & $0.00_{0.00}$  & $0.00_{0.00}$  & $0.00_{0.00}$ & $0.00_{0.00}$  & $0.00$  \\ 
    CrossFire       & \ding{51} & $0.00_{0.00}$  & $0.00_{0.00}$ & $0.00_{0.00}$  & $0.00_{0.00}$  & $0.00_{0.00}$ & $0.00_{0.00}$ & $0.00$  \\ 
    Illusion Attack  & \ding{51} & $15.98_{0.33}$  & $2.53_{0.80}$  & $45.06_{0.18}$  & $10.39_{0.11}$  & $5.47_{1.72}$  & $9.59_{0.18}$  & $14.84$ \\ 
    \cellcolor{Gray}\textbf{PTA (Ours)} & \cellcolor{Gray}\ding{51} & \cellcolor{Gray}$\textbf{74.94}_{2.67}$  & \cellcolor{Gray}$\textbf{76.64}_{0.20}$  & \cellcolor{Gray}$\textbf{76.22}_{0.24}$  & \cellcolor{Gray}$\textbf{50.11}_{0.64}$  & \cellcolor{Gray}$\textbf{64.75}_{0.35}$  & \cellcolor{Gray}$\textbf{28.13}_{0.09}$  & \cellcolor{Gray}$\textbf{61.80}$ \\ 

    \bottomrule[1.1pt]
  \end{tabular}
  }
  }
  \vspace{-5mm}
\end{table*}

\section{Experiments}

\noindent\textbf{Overview.} 
In this section, we first introduce our experimental setup. 
Then, we present a comparative analysis of our method against baseline approaches, focusing on the undetectability and generalizability of different adversarial attacks.
Next, we discuss the effectiveness of PTA in more challenging scenarios (black-box attacks, textual or audio AEs, and potential defenses beyond anomaly detection).
Finally, we analyze hyperparameter factors that could potentially affect our PTA.

\subsection{Experimental Settings}
\label{sec:effectiveness_setting}
\noindent\textbf{Models, downstream tasks and datasets.} We use three recent multimodal pre-trained models: ImageBind \cite{girdhar2023imagebind}, LanguageBind \cite{zhu2024languagebind}, and One-PEACE \cite{wang2023onepeace}. ImageBind and LanguageBind support six modalities, while One-PEACE handles three modalities: image, text, and audio.
As for downstream tasks, our experiments encompass two cross-modal matching tasks: classification and retrieval. 
For classification, we evaluate on ImageNet~\cite{deng2009imagenet} and XmediaNet~\cite{peng2018overview}, where the adversary knows target categories but not exact prompts, and target-modal proxies are generated via handcrafted templates and LLM descriptions, with same-category images as source-modal proxies. 
For retrieval, we evaluate on MSCOCO~\cite{lin2014microsoft} and XmediaNet, where the adversary knows query keywords but not exact formulations, and captions containing these keywords serve as target-modal proxies, with corresponding images as source-modal proxies.
For these tasks, we generate AEs from the image modality, with true targets located in the text modality. 
{\color{highlightcolor}We also evaluate situations where the AEs are text or audio.}

\noindent\textbf{Compared baselines.} 
We compare PTA with prevailing targeted and untargeted attacks on multimodal pre-trained models. 
For \textit{targeted attacks}, in addition to Illusion Attack \cite{zhang2024adversarial}, we also incorporate CrossFire \cite{dou2024adversarialattacksmultimodalmodels} and MF-ii \cite{zhao2024evaluating}, which optimize AEs with the source-modal example generated based on the cross-modal target.
For \textit{untargeted attacks}, we compare PTA with Sep-Attack \cite{madry2019, li-etal-2020-bert-attack}, Co-Attack \cite{zhang2022adversariala}, SGA \cite{lu2023setlevel}, and CMI-Attack \cite{fu2024improvingadversarialtransferabilityvisionlanguage} from the perspective of poisoned retrieval system performance degradation.

\noindent\textbf{Metrics.} 
For classification, we assess the adversarial attacks using the Classification Attack Success Rate (\textit{Cls ASR}), which quantifies the percentage of AEs successfully classified as the target class. 
In retrieval, we evaluate performance using the Recall at Rank K Attack Success Rate (\textit{R@K ASR}), measuring the proportion of AEs retrieved within the top-K results that match the true target. 
{\color{highlightcolor}With anomaly detection methods \cite{angiulli2002fastknn,breunig2000lof,liu2008isolation,hoffmann2007kernelpca} enabled, we report \textit{Cls ASRD} and \textit{R@K ASRD}, i.e., the fraction of all attack attempts that both bypass detection and achieve the target objective.}
We evaluate the effectiveness of PTA along two axes:
\begin{itemize}[left=0pt]
  \item To evaluate the generalizability, we test the performance (\textit{Cls ASR} or \textit{R@K ASR}) of AEs in zero-shot classification (text as prompt) and text-to-image retrieval tasks. 
  \item {\color{highlightcolor}To measure the undetectability, we apply traditional anomaly detection methods. We assess the performance (\textit{Cls ASRD} or \textit{R@K ASRD}) by counting AEs that both bypass detection and succeed, while keeping all attack attempts in the denominator.}
\end{itemize}

\noindent\textbf{Hyperparameters.} 
We use PGD attack~\cite{madry2019} under the $L_{\infty}$-norm with 100 iterations and $\epsilon=8/255$. 
For anomaly detection defense, we set $K=50$ for top-$K$ samples with filtering ratio $r = 1 - \frac{N_{\text{adv}}}{K}$. 
The detection iterations are set to $T=2$.
For PTA, the number of source-modal proxies $N_s$ is 20 (retrieval) or 25 (classification), and target-modal proxies $N_c$ is 50 (retrieval) or 10 (classification). 
The balancing factor $\alpha$ is set to 0.4 (retrieval) or 1.0 (classification) when defense is applied, and 0 otherwise to prioritize generalizability.

\subsection{The Effectiveness of PTA}
\label{sec:effectiveness_pta}
In this part, we evaluate the generalizability and undetectability of AEs generated by PTA in a white-box setting.
To evaluate the \textit{generalizability}, we optimize AEs knowing the targeted estimated distribution $\mathcal{P}_{\text{target}}(\mathbf{Y} \sim \mathcal{D}_{\text{T}} \vert Q)$ but without complete details about the true targets. 
We select two disjoint subsets from the distribution to serve as proxy targets and true targets, allowing us to assess the generalizability performance of the AEs. 
Also, we evaluate the \textit{undetectability} of AEs by applying our anomaly detection tailored for multimodal embeddings.

\noindent\textbf{Classification task.}
As shown in \cref{tab:classification_whitebox}, PTA surpasses the adopted baselines by a large margin. 
We also observe that multimodal classification systems are more vulnerable than retrieval systems (shown in \cref{tab:retrieval_whitebox}). 
We conjecture that this is because the target distribution in classification (i.e., class prompts) is more concentrated than in retrieval (i.e., user queries), making alignment feasible even for less generalizable AEs. 
To verify this, we compute the mean cosine similarity among samples drawn from the estimated source/target distributions and the trace of their covariance matrices for both tasks. 
As shown in \cref{tab:onepeace_diagnostic}, retrieval exhibits markedly lower mean cosine similarity and higher covariance traces, indicating a more dispersed distribution. 
This explains why classification systems are more vulnerable to generalized AEs, and suggests that increasing the diversity of class prompts could potentially improve adversarial robustness.

\begin{table}[!htbp]
  \centering
  \small
  \caption{
Comparison of \textbf{text-to-image retrieval degradation} by injecting varying fractions of AEs. 
\textit{Injection Ratio} is the proportion of AEs to all images. 
Results are reported as \textit{R@1 (\%) after injection (\(\downarrow\) drop in R@1 (\%))}. 
Here, \emph{R@1} (\emph{Recall@1}) is the fraction of queries whose top-ranked result is its corresponding ground-truth image.
\textbf{Lower R@1 indicates stronger disruption brought by AEs.}
}
\label{tab:inject_ae}
\resizebox{1.00\linewidth}{!}{
\setlength{\tabcolsep}{0.5mm}{
\begin{tabular}{cccc}
\toprule[1.1pt]
\textbf{Attack (Injection ratio)} & \textbf{ImageBind}       & \textbf{LanguageBind}    & \textbf{One-PEACE}      \\ 
\cmidrule(lr){1-1}\cmidrule(lr){2-4}
\underline{No Attack} (0)           & $\mathit{41.02}$                    & $\mathit{39.62}$                    & $\mathit{37.47}$                   \\ 
Sep-Attack ($10\%$)       & $38.54 \, (\downarrow 2.48)$           & $37.38 \, (\downarrow 2.24)$           & $35.73 \, (\downarrow 1.74)$          \\
Co-Attack ($10\%$)        & $37.34 \, (\downarrow 3.68)$           & $35.69 \, (\downarrow 3.93)$           & $34.31 \, (\downarrow 3.16)$          \\
SGA ($10\%$)              & $36.86 \, (\downarrow 4.16)$           & $35.54 \, (\downarrow 4.08)$           & $33.91 \, (\downarrow 3.56)$          \\ 
CMI-Attack ($10\%$)              & $ 36.90 \, (\downarrow 4.12)$           & $ 36.00 \, (\downarrow 3.62)$           & $ \, 33.80(\downarrow 3.67)$          \\ 
Illusion Attack ($1\%$)           & $37.42 \, (\downarrow 3.60)$           & $34.55 \, (\downarrow 5.07)$           & $36.55 \, (\downarrow 0.92)$          \\
\cellcolor{Gray}\textbf{Our PTA ($0.1\%$)} & \cellcolor{Gray}$33.27 \, (\downarrow 7.75)$           & \cellcolor{Gray}$36.23 \, (\downarrow 3.39)$           & \cellcolor{Gray}$37.26 \, (\downarrow 0.21)$          \\ 
\cellcolor{Gray}\textbf{Our PTA ($0.5\%$)} & \cellcolor{Gray}$23.29 \, (\downarrow 17.73)$          & \cellcolor{Gray}$15.47 \, (\downarrow 24.15)$          & \cellcolor{Gray}$33.55 \, (\downarrow 3.09)$          \\ 
\cellcolor{Gray}\textbf{Our PTA ($1\%$)} & \cellcolor{Gray}$\textbf{20.04} \, (\downarrow \textbf{20.98})$ & \cellcolor{Gray}$\textbf{12.26} \, (\downarrow \textbf{27.36})$ & \cellcolor{Gray}$\textbf{32.83} \, (\downarrow \textbf{4.64})$ \\ 
\bottomrule[1.1pt]
\end{tabular}
  }
  }
\end{table}

\begin{table}[!t]
  \centering
  \color{highlightcolor}
  \caption{Model-specific embedding geometry on MSCOCO retrieval and XmediaNet classification.}
  \label{tab:onepeace_diagnostic}
  \resizebox{0.99\linewidth}{!}{
  \setlength{\tabcolsep}{1mm}{
\begin{tabular}{llcc}
\toprule[1.1pt]
\textbf{Metric} & \textbf{Model} & \textbf{Retrieval} & \textbf{Classification} \\
\midrule
\multirow{3}{*}{\shortstack[l]{Trace\\of Cov.}}
& ImageBind    & 0.4341 / 0.4020 & 0.1917 / 0.2680 \\
& LanguageBind & 0.3711 / 0.3691 & 0.1537 / 0.2750 \\
& One-PEACE    & 0.4793 / 0.4358 & 0.2979 / 0.3062 \\
\midrule
\multirow{3}{*}{\shortstack[l]{Cosine\\Sim.}}
& ImageBind    & 0.5571 / 0.5898 & 0.7870 / 0.7208 \\
& LanguageBind & 0.6213 / 0.6234 & 0.8292 / 0.7136 \\
& One-PEACE    & 0.5109 / 0.5553 & 0.6690 / 0.6810 \\
\bottomrule[1.1pt]
\end{tabular}
  }}
  \vspace{-5mm}
  \end{table}

\begin{table*}[!htbp]
  \centering
  \small
  \caption{Comparison results of \textbf{generalizability} of AEs in \textbf{black-box attacks}. Results are reported for different tasks (\textit{Cls ASR (\%)} for classification and \textit{R@1 ASR (\%)} for retrieval).}
  \label{tab:black_box_attack_performance}
\resizebox{0.99\textwidth}{!}{
\setlength{\tabcolsep}{0.3mm}{
  \begin{tabular}{l ccc cc cc cc}
    \toprule[1.1pt]
    & & \multicolumn{4}{c}{\textbf{Classification}} & \multicolumn{4}{c}{\textbf{Retrieval}} \\
    \cmidrule(lr){3-6} \cmidrule(lr){7-10}
    & & \multicolumn{2}{c}{\textbf{XmediaNet}} & \multicolumn{2}{c}{\textbf{ImageNet}} & \multicolumn{2}{c}{\textbf{XmediaNet}} & \multicolumn{2}{c}{\textbf{MSCOCO}} \\
    \cmidrule(lr){3-4} \cmidrule(lr){5-6} \cmidrule(lr){7-8} \cmidrule(lr){9-10}
    \textbf{Attack} & \textbf{Queries} & ImageBind & LanguageBind & ImageBind & LanguageBind & ImageBind & LanguageBind & ImageBind & LanguageBind \\ 
    \cmidrule(r){1-2} \cmidrule(lr){3-4} \cmidrule(lr){5-6} \cmidrule(lr){7-8} \cmidrule(lr){9-10}
    Illusion Attack & $10^4$ & $49.58_{1.77}$  & $65.00_{0.00}$  & $33.11_{1.27}$  & $49.51_{0.93}$  & $36.96_{0.74}$  & $39.57_{1.72}$  & $7.86_{0.11}$  & $8.22_{0.53}$  \\ 
    \cellcolor{Gray}\textbf{PTA (Ours)} & \cellcolor{Gray}$10^4$ & \cellcolor{Gray}$\textbf{51.67}_{1.18}$  & \cellcolor{Gray}$\textbf{68.75}_{2.95}$  & \cellcolor{Gray}$\textbf{34.75}_{0.89}$  & \cellcolor{Gray}$\textbf{52.44}_{0.88}$  & \cellcolor{Gray}$\textbf{60.72}_{5.10}$  & \cellcolor{Gray}$\textbf{61.74}_{1.39}$  & \cellcolor{Gray}$\textbf{20.77}_{1.43}$  & \cellcolor{Gray}$\textbf{27.63}_{2.80}$  \\ 
    \cmidrule(r){1-2} \cmidrule(lr){3-4} \cmidrule(lr){5-6} \cmidrule(lr){7-8} \cmidrule(lr){9-10}
    Illusion Attack & $2\cdot10^4$ & $75.00_{0.00}$  & $87.08_{0.59}$  & $59.87_{1.35}$  & $75.96_{0.02}$  & $61.35_{0.00}$  & $60.93_{0.00}$  & $11.56_{0.00}$  & $11.64_{0.64}$  \\ 
    \cellcolor{Gray}\textbf{PTA (Ours)} & \cellcolor{Gray}$2\cdot10^4$ & \cellcolor{Gray}$\textbf{78.33}_{0.00}$  & \cellcolor{Gray}$\textbf{91.67}_{0.00}$  & \cellcolor{Gray}$\textbf{64.32}_{3.54}$  & \cellcolor{Gray}$\textbf{81.75}_{1.45}$  & \cellcolor{Gray}$\textbf{84.72}_{0.00}$  & \cellcolor{Gray}$\textbf{83.00}_{0.00}$  & \cellcolor{Gray}$\textbf{41.81}_{0.00}$  & \cellcolor{Gray}$\textbf{50.14}_{1.22}$  \\ 
    \bottomrule[1.1pt]
    \end{tabular}
  }
  }
\end{table*}

\begin{table*}[!t]
  \centering
  \color{highlightcolor}
  \caption{Cross-model transfer attacks on ImageNet zero-shot image classification. Results are reported in Cls ASR (\%).}
  \label{tab:transfer_attack}
  \resizebox{0.99\linewidth}{!}{
  \setlength{\tabcolsep}{6mm}{
  \begin{tabular}{lcccc}
  \toprule[1.1pt]
  \multirow{2}{*}{\textbf{Surrogate Model}} & \multicolumn{4}{c}{\textbf{Target Model}} \\
  \cmidrule(lr){2-5}
  & \textbf{ImageBind} & \textbf{OpenCLIP-ViT} & \textbf{OpenCLIP-RN50} & \textbf{AudioCLIP} \\
  \midrule
  ImageBind & -- & 28.65$\rightarrow$\textbf{39.19} & 28.65$\rightarrow$\textbf{41.89} & 32.97$\rightarrow$\textbf{40.81} \\
  OpenCLIP-ViT & 30.27$\rightarrow$\textbf{30.81} & -- & 30.81$\rightarrow$\textbf{39.19} & 34.59$\rightarrow$\textbf{42.43} \\
  OpenCLIP-RN50 & 27.57$\rightarrow$\textbf{37.03} & 26.49$\rightarrow$\textbf{34.86} & -- & 31.35$\rightarrow$\textbf{43.51} \\
  AudioCLIP & 27.57$\rightarrow$\textbf{38.11} & 25.41$\rightarrow$\textbf{37.57} & 32.43$\rightarrow$\textbf{41.35} & -- \\
  \bottomrule[1.1pt]
  \end{tabular}
  }}
  \vspace{-5mm}
  \end{table*}

\noindent\textbf{Retrieval task.}
Shown in \cref{tab:retrieval_whitebox}, our approach substantially improves both ASR and ASR under anomaly detection (ASRD) over baselines \cite{zhang2024adversarial, dou2024adversarialattacksmultimodalmodels, zhao2024evaluating}. 
This is because using multiple cross-modal proxy targets enhances AE generalizability to semantically similar texts, while incorporating source-proxy targets tightens alignment with the source modality and improves embedding stealthiness. 
Further, we quantify the risk of highly generalizable AEs in retrieval systems by considering injecting AEs into the image gallery as poison to compromise the \textit{overall} retrieval performance on MSCOCO \cite{lin2014microsoft}.
Specifically, different from \cref{tab:retrieval_whitebox}, we test the system with \textit{all} queries in examples of MSCOCO.
Unlike untargeted AEs, which typically break only the link between their single query, generalized targeted AEs attract many more semantically related queries.
{\color{highlightcolor}As a result, PTA causes larger performance degradation with fewer injected AEs (\cref{tab:inject_ae}) than four recent untargeted attacks \cite{madry2019, li-etal-2020-bert-attack,zhang2022adversariala,lu2023setlevel,fu2024improvingadversarialtransferabilityvisionlanguage} and Illusion Attack \cite{zhang2024adversarial}, demonstrating its high attack effectiveness and efficiency.}

{\color{highlightcolor}
{\color{highlightcolor2}Interestingly, we observe a model-specific phenomenon: One-PEACE is harder to attack in MSCOCO retrieval but gives Illusion Attack a higher ASRD under anomaly detection.}
This phenomenon is consistent with the embedding geometry in \cref{tab:onepeace_diagnostic}.
\begin{itemize}[left=0pt]
  \item \textit{Retrieval without anomaly detection.}
  {\color{highlightcolor2}On MSCOCO retrieval, PTA obtains lower absolute R@1 ASR on One-PEACE ($30.69\%$) than on ImageBind ($71.31\%$) and LanguageBind ($87.09\%$), and Illusion Attack shows the same model-specific difficulty ($9.59\%$ on One-PEACE versus $20.33\%$ and $29.41\%$ on ImageBind and LanguageBind).}
  \Cref{tab:onepeace_diagnostic} shows that One-PEACE has larger source/target covariance traces and lower within-modality mean cosine similarities, indicating a more dispersed embedding space.
  These diagnostics measure within-modality dispersion rather than directly isolating the source--target modality gap, but they suggest a harder model-specific geometry for proxy-based target generalization.
  According to Theorem~\ref{theo_1}, larger source/target variances make the generalizability--undetectability trade-off harder, which explains the lower absolute retrieval ASR on One-PEACE.
  \item \textit{Classification with anomaly detection.}
  In XmediaNet classification, the detector depends strongly on the source-modal embedding distribution.
  For Illusion Attack, classification is easier to optimize than retrieval because the target prompts are more concentrated, so the defended classification result is more affected by whether the detector separates the AE from benign source samples.
  {\color{highlightcolor2}The broader One-PEACE source-modal distribution makes target-aligned AEs less separable from benign samples, explaining why Illusion Attack keeps a higher Cls ASRD on One-PEACE ($66.25\%$) than on ImageBind and LanguageBind ($18.33\%$).
  The same dispersed geometry also makes PTA's ASR--ASRD trade-off harder, but PTA still improves One-PEACE Cls ASRD from $66.25\%$ to $87.92\%$.}
\end{itemize}
}

\subsection{\color{highlightcolor}PTA under Challenging Conditions}
\label{sec:effectiveness_pta_tc}

Here, we assess PTA's performance in tougher conditions: (i) in black-box settings, (ii) with textual or audio AEs, and (iii) against defenses of adversarial training, data augmentation, or adversarial purification. 
\textit{Since source-modal target optimization methods \cite{dou2024adversarialattacksmultimodalmodels, zhao2024evaluating} are much weaker in the white-box results, this section focuses on the stronger Illusion Attack baseline.}

\subsubsection{Black-box attacks.}
\label{blackbox}
{\color{highlightcolor}
Black-box threat models assume that adversaries cannot access the internal parameters or gradients of the target encoder.
We evaluate PTA under two common black-box settings: query-based black-box attacks and transfer-based black-box attacks.
}

{\color{highlightcolor}
\noindent\textbf{Query-based black-box attacks.}
In the query-based setting, the attacker can only query the target encoder.
Following prior black-box attack settings, we evaluate PTA using the gradient-free Square Attack~\cite{andriushchenko2020square}.
All attacks are conducted under the $L_{\infty}$-norm constraint with a maximum perturbation budget of $\epsilon=16/255$.
PTA and Illusion Attack use the same query budgets.
As reported in \cref{tab:black_box_attack_performance}, PTA consistently outperforms Illusion Attack across both classification and retrieval tasks.
These results show that PTA remains effective even when exact gradients are unavailable.
}

{\color{highlightcolor}
\noindent\textbf{Transfer-based black-box attacks.}
We also conduct cross-model transfer attacks on ImageNet zero-shot image classification using ImageBind, OpenCLIP-ViT \cite{ilharco_gabriel_2021_5143773}, OpenCLIP-RN50 \cite{ilharco_gabriel_2021_5143773}, and AudioCLIP \cite{guzhov2022audioclip} as surrogate and target encoders.
Each entry in \cref{tab:transfer_attack} is reported as Illusion Attack $\rightarrow$ PTA, where image adversarial examples are generated on the surrogate encoder under $\epsilon=8/255$ and directly evaluated on the target encoder.
As shown in \cref{tab:transfer_attack}, PTA improves over Illusion Attack across the tested surrogate--target pairs, indicating that the proxy-based objective also improves transfer-based black-box attacks.
}

{\color{highlightcolor}
Overall, PTA improves over Illusion Attack under both query-based and transfer-based black-box threat models.
}

\begin{table}[!t]
  \centering
  \caption{Comparison results of \textbf{generalizability and undetectability} of \textbf{textual or audio} AEs. We report the performance of AEs by \textit{Cls ASR (\%)} (\textit{Cls ASRD (\%)} with defense) and \textit{R@1 ASR (\%)} (\textit{R@1 ASRD (\%)} with defense) for classification and retrieval.}
  \label{tab:more_source_modality}
  \setlength{\tabcolsep}{2pt}
  \resizebox{\linewidth}{!}{%
    \begin{tabular}{ll cc cc}
      \toprule[1.1pt]
      \multirow{2}{*}{\textbf{Task}} & \multirow{2}{*}{\textbf{Method}} & \multicolumn{2}{c}{\textbf{Text Modality}} & \multicolumn{2}{c}{\textbf{Audio Modality}} \\
      \cmidrule(lr){3-4} \cmidrule(lr){5-6} 
      & & Defense \ding{55} & Defense \ding{51}& Defense \ding{55} & Defense \ding{51}\\ 
      \midrule
      \multirow{2}{*}{CLS} 
      & Illusion Attack & $26.01_{0.35}$ & $10.54_{0.21}$ & $\mathbf{100.00}_{0.00}$ & $21.45_{0.55}$ \\
      & \cellcolor{Gray}\textbf{PTA (Ours)} & \cellcolor{Gray}$\mathbf{37.32}_{0.29}$ & \cellcolor{Gray}$\mathbf{25.88}_{0.40}$ & \cellcolor{Gray}$\mathbf{100.00}_{0.00}$ & \cellcolor{Gray}$\mathbf{91.03}_{0.78}$ \\ 
      \midrule
      \multirow{2}{*}{RET} 
      & Illusion Attack & $10.67_{0.03}$ & $1.35_{0.15}$ & $0.26_{0.04}$ & $0.05_{0.01}$ \\
      & \cellcolor{Gray}\textbf{PTA (Ours)} & \cellcolor{Gray}$\mathbf{24.33}_{0.27}$ & \cellcolor{Gray}$\mathbf{18.91}_{0.32}$ & \cellcolor{Gray}$\mathbf{65.33}_{0.17}$ & \cellcolor{Gray}$\mathbf{48.17}_{0.25}$ \\ 
      \bottomrule[1.1pt]
    \end{tabular}%
  }
  \vspace{-5mm}
\end{table}

\begin{table*}[!htbp]
  \centering
  \small
  \caption{
  \textit{R@1 ASR (\%)} under varying amounts of prior knowledge (number of known entity classes) for Illusion Attack vs.\ PTA on MSCOCO.}
  \label{tab:targeted_entities_attack_performance}
  \resizebox{1.0\linewidth}{!}{
  \setlength{\tabcolsep}{8mm}{
  \begin{tabular}{llccc}
    \toprule[1.1pt]
    \textbf{Known Ent.} & \textbf{Attack} & \textbf{ImageBind} & \textbf{LanguageBind} & \textbf{One-PEACE} \\
    \midrule
    \multirow{2}{*}{``boat''} 
    & Illusion Attack      & $12.50_{0.88}$  & $17.38_{0.53}$  & $6.38_{0.23}$  \\
    & \cellcolor{Gray}\textbf{PTA (Ours)}           & \cellcolor{Gray}$\mathbf{69.94}_{0.97}$  & \cellcolor{Gray}$\mathbf{89.04}_{2.21}$  & \cellcolor{Gray}$\mathbf{23.31}_{0.09}$ \\
    \midrule
    \multirow{2}{*}{``boat'', ``person''} 
    & Illusion Attack      & $12.88_{0.88}$  & $21.75_{0.00}$  & $6.06_{0.43}$  \\
    & \cellcolor{Gray}\textbf{PTA (Ours)}           & \cellcolor{Gray}$\mathbf{68.62}_{0.88}$  & \cellcolor{Gray}$\mathbf{88.25}_{1.77}$  & \cellcolor{Gray}$\mathbf{25.38}_{0.53}$  \\
    \midrule
    \multirow{2}{*}{\makecell{``boat'', ``person'', \\``car''}} 
    & Illusion Attack      & $34.94_{1.86}$  & $49.00_{1.52}$  & $19.88_{0.35}$  \\
    & \cellcolor{Gray}\textbf{PTA (Ours)}           & \cellcolor{Gray}$\mathbf{80.44}_{0.80}$  & \cellcolor{Gray}$\mathbf{88.00}_{0.52}$  & \cellcolor{Gray}$\mathbf{53.12}_{0.33}$  \\
    \bottomrule[1.1pt]
  \end{tabular}
  }}
  \vspace{-2mm}
\end{table*}

\begin{table*}[!t]
  \centering
  \small
  \caption{Comparison results of \textbf{unknown-modal generalizability}. We report audio modality attack success rates (\textit{Cls aud ASR (\%)} and \textit{R@1 aud ASR (\%)}) on XmediaNet.}
    \label{tab:cross_modality_generalization}
\resizebox{1.00\linewidth}{!}{
\setlength{\tabcolsep}{8mm}{
  \begin{tabular}{llccc}
    \toprule[1.1pt]
    \textbf{Task} & \textbf{Method} & \textbf{ImageBind} & \textbf{LanguageBind} & \textbf{One-PEACE} \\
    \midrule
    \multirow{2}{*}{Classification} 
    & Illusion Attack      & $8.78_{0.00}$  & $13.13_{0.00}$  & $13.67_{0.00}$  \\
    & \cellcolor{Gray}\textbf{PTA (Ours)}   & \cellcolor{Gray}$\textbf{11.08}_{0.06}$  & \cellcolor{Gray}$\textbf{17.00}_{0.39}$  & \cellcolor{Gray}$\textbf{17.64}_{0.05}$  \\
        \midrule
    \multirow{2}{*}{Retrieval} 
    & Illusion Attack      & $9.11_{0.00}$  & $4.55_{0.00}$  & $13.65_{0.00}$  \\
    & \cellcolor{Gray}\textbf{PTA (Ours)}   & \cellcolor{Gray}$\textbf{9.33}_{0.11}$  & \cellcolor{Gray}$\textbf{39.75}_{0.44}$  & \cellcolor{Gray}$\textbf{19.65}_{0.07}$  \\
    \bottomrule[1.1pt]
  \end{tabular}
  }
  }
  \vspace{-2mm}
  
\end{table*}

\begin{table*}[!htbp]
  \centering
  \small
  \caption{Impact of adversarially fine-tuned CLIP VIT/B-32 on \textit{Cls ASR (\%)} and \textit{R@1 ASR (\%)} performance of attacks for classification and retrieval tasks.
  }
  \label{tab:adversarial_training_eps}
  \resizebox{1.0\linewidth}{!}{  
  \setlength{\tabcolsep}{8mm}{
  \begin{tabular}{l ccc ccc c}
    \toprule[1.1pt]
    & & \multicolumn{2}{c}{\textbf{Classification Task}} & \multicolumn{2}{c}{\textbf{Retrieval Task}} \\
    \cmidrule(lr){3-4} \cmidrule(lr){5-6}
    \textbf{Attack} & $\epsilon$ & \textbf{XmediaNet} & \textbf{ImageNet} & \textbf{XmediaNet} & \textbf{MSCOCO} \\
    \cmidrule(r){1-2} \cmidrule(lr){3-4} \cmidrule(lr){5-6}
    Illusion Attack  & $8/255$  & $22.08_{0.59}$ & $10.57_{0.13}$ & $9.72_{0.36}$  & $3.94_{0.04}$  \\ 
    \cellcolor{Gray}\textbf{PTA (Ours)}     & \cellcolor{Gray}$8/255$  & $\cellcolor{Gray}\textbf{22.50}_{0.00}$ & $\cellcolor{Gray}\textbf{13.73}_{0.05}$ & $\cellcolor{Gray}\textbf{14.20}_{0.35}$ & $\cellcolor{Gray}\textbf{4.64}_{0.07}$  \\ 
    \midrule
    Illusion Attack  & $16/255$ & ${\color{highlightcolor2}\textbf{80.42}}_{0.59}$ & $44.74_{0.01}$ & $62.31_{0.24}$ & $10.89_{0.02}$  \\
    \cellcolor{Gray}\textbf{PTA (Ours)}     & \cellcolor{Gray}$16/255$ & $\cellcolor{Gray}{\color{highlightcolor2}78.16}_{0.00}$ & $\cellcolor{Gray}\textbf{51.56}_{0.02}$ & $\cellcolor{Gray}\textbf{78.03}_{0.20}$ & $\cellcolor{Gray}\textbf{24.30}_{0.15}$  \\
    \midrule
    Illusion Attack  & $32/255$ & $99.58_{0.59}$ & $68.72_{0.09}$ & $87.52_{0.00}$ & $22.55_{0.29}$  \\ 
    \cellcolor{Gray}\textbf{PTA (Ours)}     & \cellcolor{Gray}$32/255$ & $\cellcolor{Gray}\textbf{100.00}_{0.00}$ & $\cellcolor{Gray}\textbf{74.00}_{0.16}$ & $\cellcolor{Gray}\textbf{97.78}_{0.08}$ & $\cellcolor{Gray}\textbf{64.42}_{0.11}$  \\ 
    \bottomrule[1.1pt]
  \end{tabular}
  }}
\end{table*}

\subsubsection{Textual or audio adversarial examples.}
Beyond image-based attacks, we further evaluate the effectiveness of our approach on adversarial examples generated from textual and audio modalities, which introduce additional optimization challenges due to their distinct data characteristics.
For textual AEs, we adopt Bert-Attack~\cite{li-etal-2020-bert-attack} with a perturbation budget limited to 10\% of tokens, and conduct evaluations on image--text retrieval and text classification tasks on MSCOCO, where images are treated as class labels in the classification setting.
For audio AEs, we apply projected gradient descent (PGD) under an $\ell_\infty$ constraint of $0.01$, and evaluate audio--text retrieval and audio classification on XmediaNet, with text prompts serving as labels for classification.

As reported in \cref{tab:more_source_modality}, adversarial attacks in the textual modality are generally less effective than their image counterparts, achieving noticeably lower attack success rates.
This observation aligns with prior findings that discrete token-level perturbations significantly constrain the adversarial search space, making it more difficult to precisely align textual embeddings with target representations in multimodal latent spaces.
Despite these inherent limitations, PTA consistently improves both the generalizability and undetectability of textual AEs across tasks, demonstrating its robustness under discrete and highly constrained perturbation regimes.

In contrast, audio AEs exhibit attack effectiveness comparable to image-based AEs, owing to the continuous nature of audio signals and the greater flexibility in adversarial optimization.
Under this setting, PTA again delivers substantial gains in both generalizability and undetectability, indicating that the benefits of proxy-based target optimization extend naturally to continuous modalities.
Overall, these results confirm that PTA generalizes well across heterogeneous source modalities, including both discrete and continuous inputs, highlighting its broad applicability in multimodal adversarial settings.

{\color{highlightcolor2}
We extend the baseline coverage to the datasets and models that are compatible with our classification/retrieval protocol.
Specifically, we evaluate AudioSet \cite{gemmeke2017audio} for audio classification and AudioCaps \cite{kim2019audiocaps} for audio--text retrieval, and include AudioCLIP \cite{guzhov2022audioclip} in addition to ImageBind when applicable.
AudioSet uses text prompts constructed from class labels, while AudioCaps is evaluated as audio-to-text retrieval.
These additional audio experiments use audio AEs with a PGD $\ell_\infty$ perturbation budget of $0.01$ and report no-defense ASR.
We do not include LLVIP in the main comparison because LLVIP is a visible-infrared fusion / pedestrian detection / image-to-image setting with bounding-box-style annotations, which is not directly compatible with our zero-shot classification and cross-modal retrieval metrics.
Thus, including LLVIP would not provide an apples-to-apples comparison for semantic target generalizability.
}

{\color{highlightcolor}
\cref{tab:additional_baseline_coverage} shows PTA improves over Illusion Attack in all added AudioSet/AudioCaps settings, with especially large gains on AudioCaps retrieval, showing consistent behavior on AudioCLIP and audio-centric benchmarks.

\begin{table}[!h]
  \centering
  \color{highlightcolor}
  \caption{Additional experiments with more audio dataset/model coverage. Results are ASR (\%).}
  \label{tab:additional_baseline_coverage}
  \resizebox{\columnwidth}{!}{
  \setlength{\tabcolsep}{2mm}{
  \begin{tabular}{llcc}
  \toprule[1.1pt]
  \multirow{2}{*}{\textbf{Model}} & \multirow{2}{*}{\textbf{Attack}} & \multicolumn{2}{c}{\textbf{Dataset / Task}} \\
  \cmidrule(lr){3-4}
  & & \textbf{AudioSet / Cls.} & \textbf{AudioCaps / Ret.} \\
  \midrule
  \multirow{2}{*}{ImageBind}
  & Illusion Attack & 8.833 & 0.000 \\
  & \cellcolor{Gray}\textbf{PTA (Ours)} & \cellcolor{Gray}9.917 & \cellcolor{Gray}65.625 \\
  \midrule
  \multirow{2}{*}{AudioCLIP}
  & Illusion Attack & 4.458 & 6.250 \\
  & \cellcolor{Gray}\textbf{PTA (Ours)} & \cellcolor{Gray}4.792 & \cellcolor{Gray}23.438 \\
  \bottomrule[1.1pt]
  \end{tabular}
  }}
  \vspace{-5mm}
  \end{table}

}

\subsubsection{Limited adversarial prior knowledge.}
In realistic retrieval scenarios, the adversary typically lacks access to the exact user query and can only infer partial semantic information, such as a small number of relevant entity keywords.
Consequently, constructing the estimated true target distribution $\mathcal{P}_{\text{target}}(\mathbf{Y}\!\sim\!\mathcal{D}_{\text{T}}\mid Q)$ critically depends on the amount of prior knowledge available to the attacker.
To systematically evaluate robustness under such uncertainty, we assess the effectiveness of PTA under varying degrees of adversarial prior knowledge by controlling the number of known entity classes in the query.
Specifically, we consider three settings on MSCOCO: one keyword (low prior), two keywords (medium prior), and three keywords (higher prior).

As shown in \cref{tab:targeted_entities_attack_performance}, PTA maintains a high attack success rate even when only a single keyword (e.g., ``boat'') is known, indicating strong generalization to unseen or partially specified targets.
Moreover, its performance improves steadily as additional prior knowledge becomes available.
In contrast, Illusion Attack exhibits consistently low success rates under one- and two-keyword settings and only achieves competitive performance when three keywords are provided.
These results demonstrate that PTA is particularly effective under limited adversarial prior knowledge, and that its advantage becomes more pronounced in realistic scenarios where exact target information is unavailable.

{
\color{highlightcolor}
{\color{highlightcolor2}Complementary limited model-access results are reported in the black-box evaluation above.}
}

\begin{figure*}[!htbp]
  \centering
  \begin{minipage}{0.48\linewidth}
      \centering
      \includegraphics[width=\linewidth]{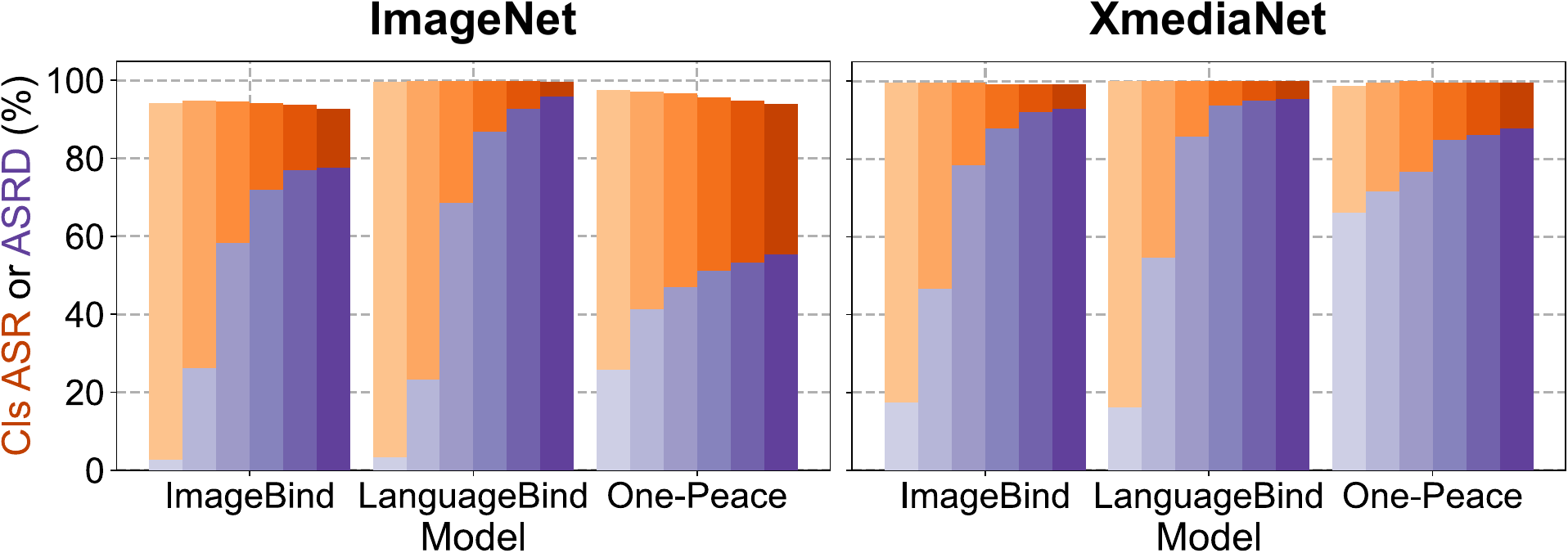}
      \caption*{Classification, $\alpha \in \{0,0.2,0.4,0.6,0.8,1.0\}$}
  \end{minipage}
  \hfill
  \begin{minipage}{0.48\linewidth}
      \centering
      \includegraphics[width=\linewidth]{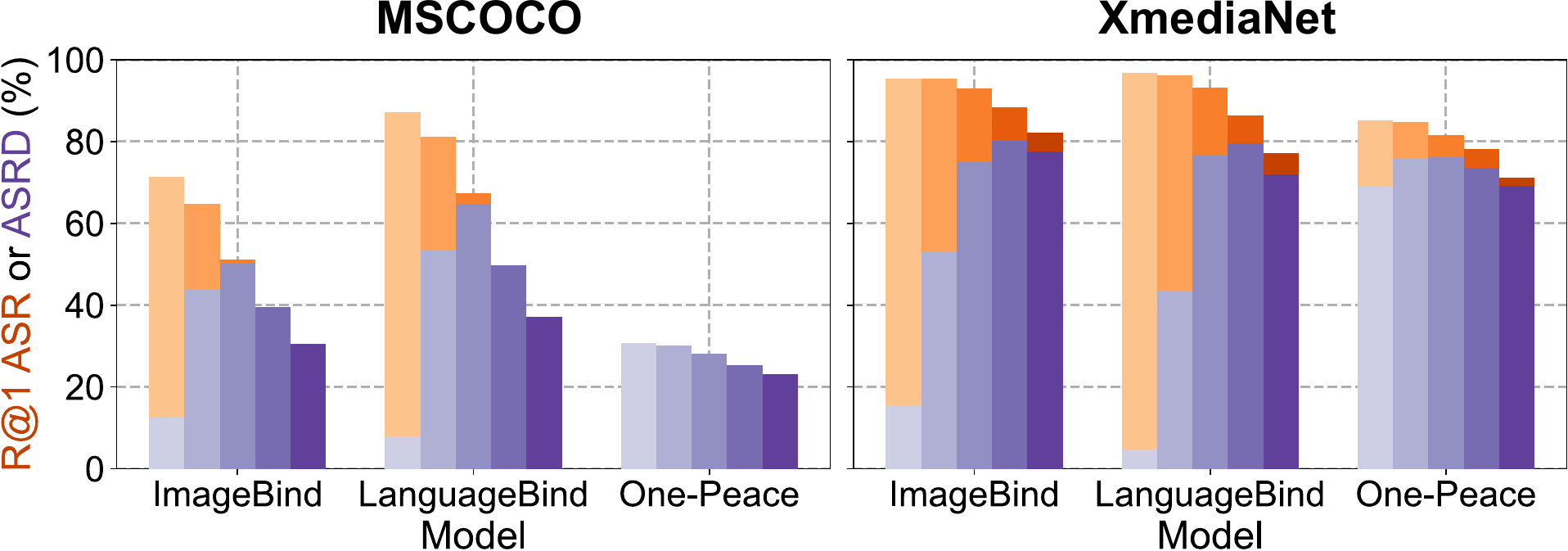}
      \caption*{Retrieval, $\alpha \in \{0,0.2,0.4,0.6,0.8\}$}
  \end{minipage}
    \caption{Analysis of the \textbf{balancing factor $\alpha$}. We present results for two metrics: ASR (Attack Success Rate (\%)) and ASRD (Attack Success Rate after anomaly Detection (\%)). ASR (\textcolor{redCustom}{red}) quantifies the generalizability of AEs, while ASRD (\textcolor{purpleCustom}{purple}) measures their undetectability.}
  \label{fig:ABL-alpha-ASR_D-and-ASR_retri}
\vspace{-3mm}
\end{figure*}

\subsubsection{PTA's effectiveness with unknown target modality.}
\label{App:Another_generalizability}
We further investigate \textit{unknown-modal generalizability}, a more challenging and realistic threat model in which the adversary has no prior knowledge of the modality of the user's input.
In this setting, adversarial examples are optimized without access to the true target modality, and success relies entirely on cross-modal semantic alignment learned by the multimodal representation.
We define generalizability to unknown-modal targets as:
\begin{align}
\nonumber
\mathbf{G}_{\text{UM}}(\mathbf{x}_{\delta}) = \mathbb{E}_{\mathbf{u} \sim \mathcal{P}_{\text{target}}^{\text{UM}}(\mathbf{U} | Q)}\left[\text{\large $\tau$}(f_{\theta_\text{S}}(\mathbf{x}_\delta), f_{\theta_\text{U}}(\mathbf{u}))\right],
\end{align}
where $f_{\theta_{\text{U}}}$ is the encoder of the unknown target modality.

To evaluate this scenario, we assume that the adversary can only access image and text modalities during optimization, while the target modality at test time is audio.
Experiments are conducted on XmediaNet.
For classification, true targets are audio samples (e.g., dog barking) used as prompts, whereas for retrieval we evaluate R@1 attack success rate in the audio-to-image retrieval setting.
This protocol explicitly prevents the attacker from exploiting modality-specific cues of the target and isolates the ability of adversarial examples to transfer across unseen modalities.

As shown in \cref{tab:cross_modality_generalization}, PTA consistently outperforms Illusion Attack under this unknown-modal setting.
This indicates that optimizing adversarial examples against a distribution of proxy targets from multiple known modalities leads to more modality-invariant perturbations.
Consequently, PTA produces adversarial examples with improved semantic alignment and enhanced transferability to unseen target modalities, highlighting its robustness under severe modality uncertainty.

\subsection{\color{highlightcolor}Evaluation Under Multiple Defenses}
\label{App:defense}
{\color{highlightcolor2}Beyond anomaly detection, we evaluate PTA under three tested non-anomaly defense families.}

\noindent\textbf{TeCoA.} {\color{highlightcolor2}For adversarial training/model hardening, we adversarially fine-tune CLIP ViT/B-32 \cite{radford2021learning} using TeCoA \cite{mao2023understanding} and report the extended classification and retrieval results in \cref{tab:adversarial_training_eps}.}
{
\color{highlightcolor2}
We extend the TeCoA adversarial fine-tuning evaluation to both classification and retrieval tasks.
Specifically, \cref{tab:adversarial_training_eps} reports results on adversarially fine-tuned CLIP ViT/B-32 for XmediaNet classification, ImageNet classification, XmediaNet retrieval, and MSCOCO retrieval under perturbation budgets of $8/255$, $16/255$, and $32/255$.
The metrics are \textit{Cls ASR (\%)} for classification and \textit{R@1 ASR (\%)} for retrieval; these are not ASRD results.
As shown in \cref{tab:adversarial_training_eps}, PTA improves most TeCoA settings and substantially improves retrieval, especially on MSCOCO.
One cell, XmediaNet classification at $\epsilon=16/255$, is slightly lower than Illusion Attack ($78.16\%$ vs.\ $80.42\%$), so we do not claim uniform superiority on every individual setting.
}

\noindent\textbf{Data augmentation.}
{\color{highlightcolor2}For input transformation defenses, following Illusion Attack \cite{zhang2024adversarial}, we use GaussianBlur, JPEG compression, and RandomAffine to disrupt adversarial features of AEs generated for LanguageBind on XmediaNet.
We optimize the adversarial noise by integrating differentiable approximations of these transformations and using Kornia \cite{shi2020differentiable} to compute gradients during backpropagation; the extended results are reported in \cref{tab:transformation_as_defense_swapped}.}
\begin{table}[h]
  \centering
  \small
  \caption{Impact of \textbf{data augmentation} on \textit{R@1 ASR (\%)} for the retrieval task on XmediaNet with LanguageBind.}
  \label{tab:transformation_as_defense_swapped}
  \resizebox{0.99\linewidth}{!}{
  \setlength{\tabcolsep}{1mm}{
  \begin{tabular}{lccc}
    \toprule[1pt]
    & \multicolumn{3}{c}{\textbf{Retrieval Task}} \\
    \cmidrule(lr){2-4}
    \textbf{Method} & \textbf{GaussianBlur} & \textbf{JPEG} & \textbf{RandomAffine} \\
    \midrule
    Illusion Attack 
      & $12.44_{0.13}$ & $9.87_{1.12}$ & $11.00_{0.48}$ \\
    \cellcolor{Gray}\textbf{PTA (Ours)}
      & \cellcolor{Gray}$\mathbf{89.33}_{0.31}$ & \cellcolor{Gray}$\mathbf{52.19}_{2.24}$ & \cellcolor{Gray}$\mathbf{63.05}_{0.16}$ \\
    \bottomrule[1pt]
  \end{tabular}
  }
  }
\end{table}

\noindent\textbf{DiffPure.}
{\color{highlightcolor2}For adversarial purification, we evaluate DiffPure \cite{nie2022diffusion} against AEs for LanguageBind on XmediaNet.
Because purification can introduce non-differentiability and stochasticity, we also test an adaptive attack using BPDA+EOT \cite{hill2020stochastic} to avoid gradient masking; the results are reported in \cref{tab:diffpure_pta}.}

\begin{table}[h]
  \centering
  \small
  \caption{Impact of \textbf{DiffPure} on R@1 ASR (\%) and cls ASR (\%) for retrieval and classification on XmediaNet with LanguageBind.}
  \label{tab:diffpure_pta}
  \resizebox{0.99\linewidth}{!}{
  \setlength{\tabcolsep}{2mm}{
  \begin{tabular}{lcc}
    \toprule[1.1pt]
    \textbf{Method} & \textbf{Classification Task} & \textbf{Retrieval Task} \\
    \midrule
    Illusion Attack & $10.31_{0.09}$ & $9.83_{0.07}$ \\
    \cellcolor{Gray}\textbf{PTA (Ours)} & \cellcolor{Gray}$\mathbf{67.62}_{0.12}$ & \cellcolor{Gray}$\mathbf{71.97}_{0.37}$ \\
    \bottomrule[1.1pt]
  \end{tabular}
  }
  }
\end{table}

{\color{highlightcolor2}
Together, \cref{tab:adversarial_training_eps}, \cref{tab:transformation_as_defense_swapped}, and \cref{tab:diffpure_pta} show that PTA remains stronger than Illusion Attack under the tested adversarial fine-tuning, input-transformation, and purification settings.
We do not claim a universal guarantee against all possible defenses; rather, these experiments validate that the PTA objective remains effective under the evaluated non-anomaly defense families.
}

\subsection{Ablation Studies}

\noindent\textbf{Balancing factor $\alpha$.}
In \cref{eq:overall_loss}, $\alpha$ is critical for controlling the alignment of AEs with the target modality versus the source modality. 
The ASR-ASRD trade-off observed in \cref{fig:ABL-alpha-ASR_D-and-ASR_retri} aligns with our theoretical analysis in \cref{sec:Trade-off_Analysis_theorem} and gives practitioners a way to precisely tune their objective: lower $\alpha$ emphasizes broad cross-target matching, whereas higher $\alpha$ emphasizes stealth. 
The effect is more pronounced for retrieval, where targets are more dispersed, but the same tuning rule holds across tasks. 
To summarize, practitioners can set a lower $\alpha$ to prioritize generalizability and a higher $\alpha$ to prioritize undetectability.

\begin{figure}
  \centering 
\includegraphics[width=0.975\linewidth]{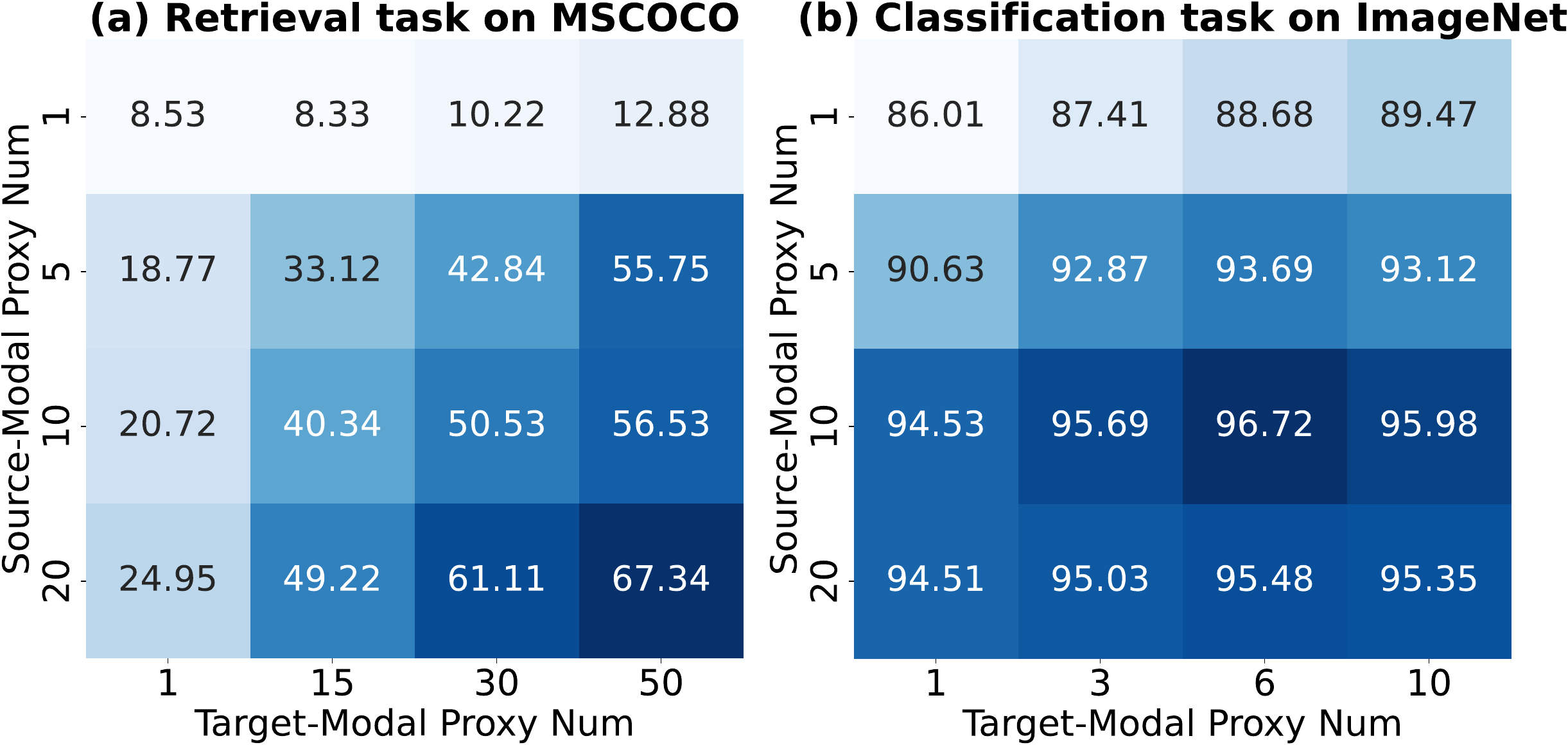}
  \caption{Attack performance with different number of \textbf{target-modal proxies} ($N_c$) and \textbf{source-modal proxies} ($N_s$).}
\label{fig:ABL-numbersOfTargets}
  \vspace{-5mm}

\end{figure}

\noindent\textbf{Number of proxy targets.} 
\cref{fig:ABL-numbersOfTargets} shows that increasing the number of target-modal or source-modal proxies can improve the attack effectiveness of AEs. 
In specific, increasing target-modal proxies markedly boosts ASR in retrieval but only modestly in classification, likely due to the lower similarity between target-modal proxies in retrieval tasks, making the training samples more versatile.
In summary, a few dozen proxies suffice for strong performance in both retrieval and classification.
We also analyze the cost of gathering and optimizing with proxies in \cref{App:additional_cost}.

\noindent\textbf{Varying prompt variety.}
We vary the diversity of $\mathcal{P}_{\mathrm{target}}(\mathbf{Y}\!\sim\!\mathcal{D}_{\mathrm{T}}\mid Q)$ by using different prompt templates for ImageNet class prompts to test PTA's performance with different classification prompts:
\begin{itemize}
  \item Standard: ``a photo of a \{class\}.’’
  \item Waffle: Prompts with random words/characters \cite{roth2023wafflinga}.
  \item Manual: 80 manually curated generic prompts \cite{radford2021learning}, e.g., ``a drawing of the \{class\}.’’
\end{itemize}
\Cref{tab:asr_performance_comparison} reports \textit{Cls ASR (\%)} for AEs trained/tested under different prompt sets. 
When test prompts are more diverse (e.g., Manual), ASR drops, supporting our conjecture and suggesting the defense of {\color{highlightcolor}increasing} prompt variety to reduce target concentration and therefore hinder the effectiveness of generalized targeted attacks.

\begin{table}[h]
  \centering
  \small
  \caption{Comparison of \textit{Cls ASR (\%)} across different train/test prompts with $\epsilon=4/255$ on ImageNet.}
  \label{tab:asr_performance_comparison}
  \resizebox{0.99\linewidth}{!}{
  \setlength{\tabcolsep}{4mm}{
  \begin{tabular}{lccc}
    \toprule[1.1pt]
    \textbf{Train \textbackslash Test} & Standard & Waffle & Manual \\
    \midrule
    Standard  & $99.58$ & $95.18$ & $80.51$ \\
    Waffle    & $99.58$ & $95.28$ & $79.95$ \\
    Manual    & $99.58$ & $95.48$ & $83.57$ \\
    \bottomrule[1.1pt]
  \end{tabular}
  }}
  \vspace{-2mm}
  
\end{table}

\subsection{Additional Cost of PTA}
\label{App:additional_cost}

\begin{table}[h]
\centering
\caption{Compute overhead of PTA vs.\ PGD (no proxies). 
Numbers are measured on ImageBind with 100 target proxies and 50 source proxies.}
\resizebox{0.99\linewidth}{!}{
\setlength{\tabcolsep}{3mm}{
\begin{tabular}{lcc}
\toprule[1.1pt]
\textbf{Method} & \textbf{Time per epoch} & \textbf{GPU memory used} \\
\midrule
PGD & 121.1 ms & 6184 MB \\
\cellcolor{Gray}\textbf{PTA (Ours)} & \cellcolor{Gray}121.3 ms ($\uparrow$ 0.16\%) & \cellcolor{Gray}6186 MB ($\uparrow$ 0.03\%) \\
\bottomrule[1.1pt]
\end{tabular}
}}
\label{tab:pta_overhead}
\end{table}
PTA introduces one extra component beyond normal adversarial attacks: a set of proxy embeddings. 
Crucially, proxy collection and embedding computation are performed \textit{offline} on high-performance machines before any attack is executed. 
At attack time, the optimization only consumes a few additional lookups/inner-products against the precomputed proxies, so the online overhead is negligible.
In specific, in \textit{classification}, AE optimization runs on the attacker’s device, while proxy collection still happens offline without time constraints. 
In \textit{retrieval}, the adversary collects proxies and optimizes adversarial examples (AEs) offline (e.g., using generative models, public datasets, or web sources), then uploads the finalized AEs to the gallery. 
All heavy computation occurs off device, so the low-resource client does not run the optimization procedure.
Since PTA only adds precomputed proxy embeddings during optimization, the extra compute/memory cost is minimal.

As shown in \cref{tab:pta_overhead}, for ImageBind on an NVIDIA V100 with 100 target proxies and 50 source proxies, PTA incurs only a \textit{0.16\%} increase in optimization time per epoch and a \textit{0.03\%} increase in GPU memory versus a normal adversarial attack (PGD). 
These differences are practically negligible, confirming that PTA remains suitable for realistic low-compute attack settings.

\subsection{\color{highlightcolor}Discussion on the collection of proxies}
\label{sec:proxy_collection_discussion}

{\color{highlightcolor}
This part discusses how proxies are constructed in our experiments and practice, how to select effective proxies, how poor proxies affect PTA, and how proxy diversity matters.
}

\subsubsection{\color{highlightcolor}Proxy construction in experiments and practice.}
\label{subsec:proxy_construction}

\begin{figure*}[!t]
  \centering
  \includegraphics[width=0.8\linewidth]{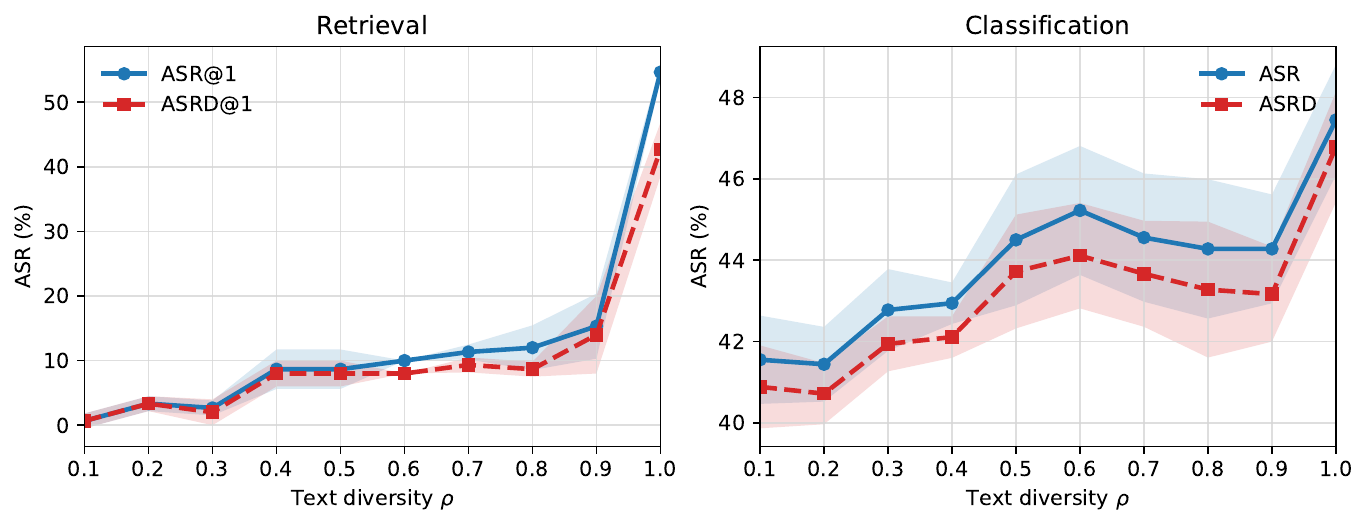}
  \caption{\color{highlightcolor}Count-fixed ablation on target-modal proxy diversity $\rho$. 
  The proxy numbers are fixed for retrieval on ImageBind/MSCOCO with $(N_s,N_c)=(20,50)$ and classification on ImageBind/ImageNet with $(N_s,N_c)=(25,10)$.
  Larger $\rho$ indicates richer contextual proxy texts.}
  \label{fig:proxy_diversity_ablation}
  \vspace{-5mm}
  \end{figure*}

{\color{highlightcolor}
In the main experiments, the adversary only knows a coarse target concept $Q$, not the exact target used at test time.
Proxy samples and true targets are split within the same dataset or category.
This controlled same-distribution proxy setting isolates generalization to unseen targets rather than additional dataset shift.
For retrieval, we collect captions/queries containing the target keyword, split them into disjoint proxy and true-query sets, and use images paired with target-modal proxy captions as source-modal proxies.
For classification, we use handcrafted prompts and LLM-generated category descriptions as target-modal proxies, and same-category images as source-modal proxies.

In practice, target-modal proxies can be collected from public datasets, web resources, or LLM-generated prompts/captions containing the target concept, while source-modal proxies can come from public datasets, web images, or generative models.
PTA therefore requires proxy samples that approximate the source-side and target-side distributions induced by $Q$, rather than the exact future query or prompt.
External, web, or generative proxies may introduce domain shift, which is why we further analyze proxy quality and diversity below.
}

\subsubsection{\color{highlightcolor}Guidelines for effective proxy selection.}
{\color{highlightcolor}
Proxy selection follows the roles of the two losses.
Target-modal proxies should approximate the possible target distribution $\mathcal{P}_{\mathrm{target}}(\mathbf{Y}\sim\mathcal{D}_{\mathrm{T}}\mid Q)$, while source-modal proxies should approximate the local benign source-modal manifold $\mathcal{P}_{\mathrm{target}}(\mathbf{X}\sim\mathcal{D}_{\mathrm{S}}\mid Q)$.
Effective target-modal proxies should be semantically relevant to $Q$ and diverse enough to cover likely target formulations, while effective source-modal proxies should be benign samples associated with the same concept and close to the local source-modal manifold.
A practical procedure is to retrieve samples related to $Q$, cluster them in the embedding space to improve coverage, remove near duplicates to avoid overfitting to one formulation, and filter outliers or noisy samples.
This guideline is consistent with Theorem~\ref{theo_1}: biased or noisy proxies can shift the empirical source/target means, enlarge the empirical modality gap, or increase source/target variance, thereby worsening the generalizability--undetectability trade-off.
}

{\color{highlightcolor2}
\linelabel{ll:c23:s}To make this procedure concrete and reproducible, we instantiate it as \textsc{ProxySelect} (\cref{alg:rev-c23-proxyselect}), a pipeline that applies three sequential cleaning steps to the keyword-retrieved target-modal candidate pool.
Greedy deduplication scans the pool and keeps a candidate only if its cosine similarity to every previously kept candidate stays below $0.95$, removing redundant targets.
Anchor-based outlier removal then drops candidates whose cosine similarity to the concept anchor falls below median${}-3\times$MAD, filtering off-concept noise.
Finally, coverage clustering runs spherical $K$-means with $K{=}N_T$ under farthest-point seeding and keeps one medoid per cluster, so the selected proxies span the target distribution.
\textsc{ProxySelect} is applied only to the target-modal candidate pool, and source-modal proxies follow directly from the selected targets.
In retrieval, the source image paired with each selected proxy caption is retained automatically.
In classification, source-modal proxies are sampled from the same-category benign images described above.
A component-wise ablation under deliberately noisy pools (\cref{tab:rev-c23-ablation}) confirms that each step substantially improves the attack performance under its matching contamination, and the full implementation will be released for reproducibility.\linelabel{ll:c23:e}
}

\begin{algorithm}[!t]
\color{highlightcolor2}
\SetAlgoLined
\DontPrintSemicolon
\KwIn{target candidate embeddings $E\!=\!\{\bar{f}_T(c)\}_{c\in\mathcal{C}_T}$ where $\bar{f}(\cdot)\!\triangleq\!f(\cdot)/\|f(\cdot)\|_2$, concept anchor $\bar{q}$, budget $N_T$}
\KwOut{selected target proxies $\mathcal{P}_T$}
\tcc{Step 1: deduplication}
Scan $E$ in order: keep $e$ iff $\cos(e,e')\!<\!0.95$ for all previously kept $e'$\;
\tcc{Step 2: outlier removal}
$s_i\!\leftarrow\!\cos(e_i,\bar{q})$ for each $e_i\!\in\!E$\;
$E\!\leftarrow\!\{e_i\!\in\!E:s_i\!\ge\!\mathrm{median}(s)\!-\!3\!\cdot\!\mathrm{MAD}(s)\}$\;
\tcc{Step 3: coverage clustering (spherical $K$-means)}
$\{\boldsymbol{c}_k\}_{k=1}^{N_T}\!\leftarrow$ farthest-point seeding on $E$ starting from its first element\;
\Repeat{assignments unchanged (at most $25$ iterations)}{
  $C_k\!\leftarrow\!\{e\!\in\!E:\arg\max_{j}\cos(e,\boldsymbol{c}_j)\!=\!k\}$ for each $k$\;
  $\boldsymbol{c}_k\!\leftarrow\!\mathrm{mean}(C_k)/\|\mathrm{mean}(C_k)\|_2$ for each $k$\;
}
$E\!\leftarrow\!\{\arg\max_{e\in C_k}\cos(e,\boldsymbol{c}_k)\}_{k=1}^{N_T}$\tcp*{medoid per cluster}
\Return $\mathcal{P}_T\!\leftarrow\!E$\;
\caption{\textsc{ProxySelect}}
\label{alg:rev-c23-proxyselect}
\end{algorithm}

\subsubsection{\color{highlightcolor}Effect of poor proxies.}
{\color{highlightcolor}
Because source-modal and target-modal proxies approximate the distributions used by $\mathcal{L}_{\mathrm{D}}$ and $\mathcal{L}_{\mathrm{G}}$, we evaluate PTA under deliberately imperfect proxy settings.

For classification, we use ImageBind on ImageNet with anomaly-detection defense enabled and report \textit{Cls ASRD (\%)}.
We compare the standard prompt setting with biased ChatGPT-style prompts (e.g., ``a high-resolution photo of a majestic African elephant standing in a grassy savanna'') and noisy Waffle-style prompts (e.g., ``a photo of an elephant, which is banana.'').
These settings increase bias or noise in the target-modal proxies.
}

\begin{table}[h]
\centering
\color{highlightcolor}
\caption{Performance of PTA with different proxy numbers in difficult classification settings. Numbers are Cls ASRD (\%). $N_s$ denotes the number of source-modal proxies and $N_c$ denotes the number of target-modal proxies.}
\label{tab:proxy_quality_classification}
\small
\setlength{\tabcolsep}{5mm}
\resizebox{\columnwidth}{!}{
\begin{tabular}{@{}lccc@{}}
\toprule[1.1pt]
\textbf{\# proxies $(N_s,N_c)$} & \textbf{Baseline} & \textbf{Biased} & \textbf{Noisy} \\
\midrule    
$(1,1)$ & 86.01 & 78.93 & 72.14 \\
$(5,3)$ & 92.87 & 88.16 & 80.45 \\
$(10,6)$ & 96.72 & 93.51 & 84.37 \\
$(20,10)$ & 95.35 & 94.72 & 86.28 \\
\bottomrule[1.1pt]
\end{tabular}
}
\end{table}

{\color{highlightcolor}
For retrieval, we use ImageBind on MSCOCO without anomaly-detection defense and report \textit{R@1 ASR (\%)}.
The baseline uses captions/queries containing the target keyword as target-modal proxies and their paired images as source-modal proxies.
The biased-proxy setting restricts target-modal proxies to a narrow contextual subset.
For example, for the target concept ``elephant'', we collect proxies only from captions containing both ``elephant'' and ``forest''.
For the distant-query setting, we first encode the disjoint true queries and source-modal proxies into the normalized shared embedding space, compute each true query's $L_2$ distance to the mean source-modal proxy embedding, and evaluate the farthest queries under the same evaluation size as the baseline.
This setting increases the proxy--target mismatch.
}

\begin{table}[h]
  \centering
  \color{highlightcolor}
  \caption{Performance of PTA with different proxy numbers in difficult retrieval settings. Numbers are R@1 ASR (\%). $N_s$ denotes the number of source-modal proxies and $N_c$ denotes the number of target-modal proxies.}
  \label{tab:proxy_quality_retrieval}
  \small
  \setlength{\tabcolsep}{2.5pt}
  \resizebox{\columnwidth}{!}{
  \begin{tabular}{@{}lccc@{}}
  \toprule[1.1pt]
  \textbf{\# proxies $(N_s,N_c)$} & \textbf{Baseline} & \textbf{Biased proxies} & \textbf{Distant queries} \\
  \midrule
  $(1,1)$ & 8.53 & 6.41 & 3.92 \\
  $(15,5)$ & 33.12 & 28.47 & 19.76 \\
  $(30,10)$ & 50.53 & 45.12 & 32.41 \\
  $(50,20)$ & 67.34 & 62.28 & 46.83 \\
  \bottomrule[1.1pt]
  \end{tabular}
  }
  \end{table}

\begin{table}[!t]
\centering
\color{highlightcolor2}
\caption{Single-target selection versus PTA on MSCOCO retrieval with ImageBind. Results are averaged over three common target concepts to ensure enough candidates.}
\label{tab:rev-c22-deliberate}
\small
\setlength{\tabcolsep}{3mm}{
\begin{tabular}{lcc}
\toprule[1.1pt]
\textbf{Method}
& \textbf{$\cos(\hat{\mathbf{z}}, \hat{\boldsymbol{\mu}})$}
& \textbf{R@1 ASR} \\
\midrule
Illusion-random                    & 0.66 & 14.06 \\
Illusion-centroid ($N$=50)         & 0.76 & 13.02 \\
Illusion-centroid ($N$=500)        & 0.78 & 17.71 \\
Illusion-centroid ($N$=5000)       & 0.79 & 18.11 \\
\textbf{PTA (Ours)}                & --   & \textbf{48.96} \\
\bottomrule[1.1pt]
\end{tabular}
}
\end{table}

\begin{table}[!t]
\centering
\color{highlightcolor2}
\caption{\textsc{ProxySelect} component ablation under noisy target-modal proxy pools.}
\label{tab:rev-c23-ablation}
\resizebox{0.8\linewidth}{!}{
\setlength{\tabcolsep}{3mm}{
\resizebox{\linewidth}{!}{%
\setlength{\tabcolsep}{0.7mm}%
\begin{tabular}{lcccc}
\toprule[1.1pt]
\multirow{2}{*}{\textbf{Noise}} & \multicolumn{2}{c}{\textbf{R@1 ASR}} & \multicolumn{2}{c}{\textbf{R@1 ASRD}} \\
\cmidrule(lr){2-3}\cmidrule(lr){4-5}
 & random & ProxySelect & random & ProxySelect \\
\midrule
Duplicate & 42.01 & \textbf{46.18} & 15.97 & \textbf{18.75} \\
Cluster   & 2.43  & \textbf{41.32} & 1.39  & \textbf{11.80} \\
Outlier   & 31.60 & \textbf{37.85} & 9.37  & \textbf{13.47} \\
\bottomrule[1.1pt]
\end{tabular}}
}}
  \vspace{-3mm}
\end{table}

{\color{highlightcolor}
Poor proxies weaken PTA by reducing target coverage or enlarging the proxy--target mismatch, while increasing proxy coverage partly mitigates the degradation.
For example, noisy classification improves from $72.14\%$ to $86.28\%$ Cls ASRD, and distant-query retrieval improves from $3.92\%$ to $46.83\%$ R@1 ASR as the amount of proxies increases.
}

\subsubsection{\color{highlightcolor}Proxy diversity with fixed proxy count.}

{\color{highlightcolor}
We conduct a count-fixed diversity ablation by keeping the numbers of source-modal and target-modal proxies unchanged and varying only the textual diversity of target-modal proxies.
For a target keyword $Q$, each target-modal proxy text $\hat{\mathbf{y}}$ is decomposed as
\[
\hat{\mathbf{y}}=[Q,u_1,\ldots,u_m],
\]
where $u_i$ are non-target contextual units such as attributes, actions, noun phrases, or prepositional phrases.
We define
\[
\mathcal{T}_{\rho}(\hat{\mathbf{y}},Q)
=
[Q,u_1,\ldots,u_{\lfloor \rho m\rfloor}],
\]
where $\rho=0$ keeps only the target concept, $\rho=0.5$ keeps partial context, and $\rho=1$ keeps the original caption or prompt.
We parse non-target contextual units using POS/chunk rules, keep the target concept $Q$, and retain $\lfloor \rho m\rfloor$ contextual units in their original order.
Thus, $\rho$ controls proxy diversity while proxy count remains fixed.
Particularly, for the classification ablation, we use a smaller adversarial budget of $\epsilon=2/255$ to avoid saturation and make the trend across $\rho$ visible.
}

{\color{highlightcolor}
\cref{fig:proxy_diversity_ablation} shows that with the proxy count fixed, richer target-modal contexts improve target-distribution coverage, especially for retrieval.
As shown in \cref{tab:onepeace_diagnostic}, retrieval targets are more dispersed than classification targets, with lower target-side mean cosine similarity and larger target-side covariance trace.
Therefore, increasing $\rho$ brings a larger gain in retrieval: R@1 ASR increases from about $1\%$ to about $55\%$, and R@1 ASRD increases from about $1\%$ to about $42\%$.
Classification targets are more concentrated, so the gain from proxy diversity is smaller: Cls ASR increases from about $42\%$ to about $48\%$, and Cls ASRD increases from about $41\%$ to about $47\%$.
}

\subsubsection{\color{highlightcolor2}Cross-domain proxy evaluation.}
{\color{highlightcolor2}
\linelabel{ll:c24:s}We additionally evaluate a cross-domain setting where proxies are drawn from a different dataset than the true targets: Flickr30k proxies$\to$MSCOCO targets for retrieval over 10 shared keywords, and Caltech-256 proxies$\to$ImageNet targets for classification over 16 shared classes.
\cref{tab:rev-c24-cross-domain} reports the results alongside same-domain controls and the Maximum Mean Discrepancy (MMD) to verify the domain shift.
Both methods mostly degrade under cross-domain shift, but PTA's drop is substantially milder, since averaging over multiple proxies mitigates the per-sample noise introduced by the domain gap.\linelabel{ll:c24:e}
}

\begin{table}[!t]
\centering
\color{highlightcolor2}
\caption{Cross-domain and same-domain proxy selection evaluation on ImageBind. The evaluation is limited to concepts shared by both datasets.}
\label{tab:rev-c24-cross-domain}
\small
\resizebox{\columnwidth}{!}{
\setlength{\tabcolsep}{2mm}{
\begin{tabular}{ll c cc}
\toprule[1.1pt]
& \textbf{Proxy} & \textbf{Shift verif.} & \multicolumn{2}{c}{\textbf{ASR / ASRD (\%)}} \\
\cmidrule(lr){3-3} \cmidrule(lr){4-5}
\textbf{Task} & \textbf{domain} & \textbf{MMD} & \textbf{Illusion Attack} & \textbf{PTA} \\
\midrule
\multirow{2}{*}{Retrieval}
  & Same  & 0.01 & 14.06 / 7.66  & 55.78 / 27.81 \\
  & Cross & 0.07 & 4.37 / 3.43   & 43.59 / 27.34 \\
\midrule
\multirow{2}{*}{Classification}
  & Same  & 0.02 & 96.61 / 88.80 & 99.48 / 94.01 \\
  & Cross & 0.17 & 87.50 / 81.77 & 99.74 / 94.01 \\
\bottomrule[1.1pt]
\end{tabular}
}}
  \vspace{-5mm}
\end{table}

\section{Conclusion}
\label{sec:conclusion}

In this paper, we investigated targeted adversarial attacks in cross-modal matching tasks by examining both undetectability and generalizability. 
Our anomaly detection analysis in the embedding space reveals that existing targeted AEs are vulnerable to detection and exhibit poor generalization to semantically similar or partially known targets.
To address these challenges, we proposed Proxy Targeted Attack (PTA), which leverages multimodal proxies to achieve both superior undetectability and generalizability. 
In addition, our theoretical findings highlight the interplay between these two limitations of AEs and demonstrate how PTA achieves an optimal balance between them.
Experiments validate PTA's effectiveness in generating undetectable AEs while maintaining a high success rate against semantically similar targets, underscoring its potential for real-world adversarial scenarios.

\ifCLASSOPTIONcaptionsoff
  \newpage
\fi

\IEEEpeerreviewmaketitle



\bibliographystyle{IEEEtran}
\bibliography{IEEEabrv,multi-label}

\begin{thebibliography}{10}
\providecommand{\url}[1]{#1}
\csname url@samestyle\endcsname
\providecommand{\newblock}{\relax}
\providecommand{\bibinfo}[2]{#2}
\providecommand{\BIBentrySTDinterwordspacing}{\spaceskip=0pt\relax}
\providecommand{\BIBentryALTinterwordstretchfactor}{4}
\providecommand{\BIBentryALTinterwordspacing}{\spaceskip=\fontdimen2\font plus
\BIBentryALTinterwordstretchfactor\fontdimen3\font minus \fontdimen4\font\relax}
\providecommand{\BIBforeignlanguage}[2]{{%
\expandafter\ifx\csname l@#1\endcsname\relax
\typeout{** WARNING: IEEEtran.bst: No hyphenation pattern has been}%
\typeout{** loaded for the language `#1'. Using the pattern for}%
\typeout{** the default language instead.}%
\else
\language=\csname l@#1\endcsname
\fi
#2}}
\providecommand{\BIBdecl}{\relax}
\BIBdecl

\bibitem{girdhar2023imagebind}
R.~Girdhar, A.~{El-Nouby}, Z.~Liu, M.~Singh, K.~V. Alwala, A.~Joulin, and I.~Misra, ``Imagebind: One embedding space to bind them all,'' in \emph{CVPR}, 2023.

\bibitem{wang2023onepeace}
P.~Wang, S.~Wang, J.~Lin, S.~Bai, X.~Zhou, J.~Zhou, X.~Wang, and C.~Zhou, ``One-peace: Exploring one general representation model toward unlimited modalities,'' \emph{arXiv preprint arXiv:2305.11172}, 2023.

\bibitem{su2023pandagpt}
Y.~Su, T.~Lan, H.~Li, J.~Xu, Y.~Wang, and D.~Cai, ``Pandagpt: One model to instruction-follow them all,'' in \emph{Proceedings of the 1st Workshop on Taming Large Language Models: Controllability in the Era of Interactive Assistants!}, D.~Hazarika, X.~R. Tang, and D.~Jin, Eds.\hskip 1em plus 0.5em minus 0.4em\relax Prague, Czech Republic: ACL, Sep. 2023, pp. 11--23.

\bibitem{xing2024seeing}
Y.~Xing, Y.~He, Z.~Tian, X.~Wang, and Q.~Chen, ``Seeing and hearing: Open-domain visual-audio generation with diffusion latent aligners,'' in \emph{CVPR}, 2024, pp. 7151--7161.

\bibitem{huang2024dreamtime}
Y.~Huang, J.~Wang, Y.~Shi, B.~Tang, X.~Qi, and L.~Zhang, ``Dreamtime: An improved optimization strategy for diffusion-guided 3d generation,'' in \emph{ICLR}, 2024.

\bibitem{li2023blip2a}
J.~Li, D.~Li, S.~Savarese, and S.~Hoi, ``Blip-2: Bootstrapping language-image pre-training with frozen image encoders and large language models,'' in \emph{ICML}, 2023.

\bibitem{jiang2024visual}
F.~Jiang, C.~Tang, L.~Dong, K.~Wang, K.~Yang, and C.~Pan, ``Visual language model based cross-modal semantic communication systems,'' \emph{arXiv preprint arXiv:2407.00020}, 2024.

\bibitem{chi2024eva}
X.~Chi, H.~Zhang, C.-K. Fan, X.~Qi, R.~Zhang, A.~Chen, C.-m. Chan, W.~Xue, W.~Luo, S.~Zhang, and Y.~Guo, ``Eva: An embodied world model for future video anticipation,'' \emph{arXiv preprint arXiv:2410.15461}, 2024.

\bibitem{lerner2024crossmodal}
P.~Lerner, O.~Ferret, and C.~Guinaudeau, ``Cross-modal retrieval for~knowledge-based visual question answering,'' in \emph{Advances in Information Retrieval}, N.~Goharian, N.~Tonellotto, Y.~He, A.~Lipani, G.~McDonald, C.~Macdonald, and I.~Ounis, Eds.\hskip 1em plus 0.5em minus 0.4em\relax Cham: Springer Nature Switzerland, 2024, pp. 421--438.

\bibitem{zhao2024evaluating}
Y.~Zhao, T.~Pang, C.~Du, X.~Yang, C.~Li, N.-M.~M. Cheung, and M.~Lin, ``On evaluating adversarial robustness of large vision-language models,'' in \emph{NeurIPS}, 2024.

\bibitem{schulhoff2023ignore}
S.~Schulhoff, J.~Pinto, A.~Khan, L.-F. Bouchard, C.~Si, S.~Anati, V.~Tagliabue, A.~Kost, C.~Carnahan, and J.~{Boyd-Graber}, ``Ignore this title and hackaprompt: Exposing systemic vulnerabilities of llms through a global prompt hacking competition,'' in \emph{EMNLP}, 2023, pp. 4945--4977.

\bibitem{fan2024unbridled}
Y.~Fan, Y.~Cao, Z.~Zhao, Z.~Liu, and S.~Li, ``Unbridled icarus: A survey of the potential perils of image inputs in multimodal large language model security,'' \emph{arXiv preprint arXiv:2404.05264}, 2024.

\bibitem{zhang2024adversarial}
T.~Zhang, R.~Jha, E.~Bagdasaryan, and V.~Shmatikov, ``Adversarial illusions in multi-modal embeddings,'' in \emph{USENIX Security}, 2024.

\bibitem{inkawhich2023adversarialattacksfoundationalvision}
N.~Inkawhich, G.~McDonald, and R.~Luley, ``Adversarial attacks on foundational vision models,'' \emph{arXiv preprint arXiv:2308.14597}, 2023.

\bibitem{dou2024adversarialattacksmultimodalmodels}
Z.~Dou, X.~Hu, H.~Yang, Z.~Liu, and M.~Fang, ``Adversarial attacks to multi-modal models,'' \emph{arXiv preprint arXiv:2409.06793}, 2024.

\bibitem{angiulli2002fastknn}
F.~Angiulli and C.~Pizzuti, ``Fast outlier detection in high dimensional spaces,'' in \emph{Principles of Data Mining and Knowledge Discovery}, T.~Elomaa, H.~Mannila, and H.~Toivonen, Eds., 2002, pp. 15--27.

\bibitem{breunig2000lof}
M.~M. Breunig, H.-P. Kriegel, R.~T. Ng, and J.~Sander, ``Lof: Identifying density-based local outliers,'' \emph{SIGMOD Rec.}, vol.~29, no.~2, pp. 93--104, 2000.

\bibitem{liu2008isolation}
F.~T. Liu, K.~M. Ting, and Z.-H. Zhou, ``Isolation forest,'' in \emph{ICDM}, Dec. 2008, pp. 413--422.

\bibitem{hoffmann2007kernelpca}
H.~Hoffmann, ``Kernel pca for novelty detection,'' \emph{Pattern Recognition}, vol.~40, no.~3, pp. 863--874, Mar. 2007.

\bibitem{xu2023multimodal}
P.~Xu, X.~Zhu, and D.~A. Clifton, ``Multimodal learning with transformers: A survey,'' \emph{TPAMI}, vol.~45, no.~10, pp. 12\,113--12\,132, 2023.

\bibitem{zong2024self}
Y.~Zong, O.~Mac~Aodha, and T.~M. Hospedales, ``Self-supervised multimodal learning: A survey,'' \emph{TPAMI}, vol.~47, no.~7, pp. 5299--5318, 2024.

\bibitem{baltruvsaitis2018multimodal}
T.~Baltru{\v{s}}aitis, C.~Ahuja, and L.-P. Morency, ``Multimodal machine learning: A survey and taxonomy,'' \emph{TPAMI}, vol.~41, no.~2, pp. 423--443, 2018.

\bibitem{zhu2024vision+}
Y.~Zhu, Y.~Wu, N.~Sebe, and Y.~Yan, ``Vision+ x: A survey on multimodal learning in the light of data,'' \emph{TPAMI}, vol.~46, no.~12, pp. 9102--9122, 2024.

\bibitem{radford2021learning}
A.~Radford, J.~W. Kim, C.~Hallacy, A.~Ramesh, G.~Goh, S.~Agarwal, G.~Sastry, A.~Askell, P.~Mishkin, J.~Clark, G.~Krueger, and I.~Sutskever, ``Learning transferable visual models from natural language supervision,'' in \emph{ICML}.\hskip 1em plus 0.5em minus 0.4em\relax PMLR, Jul. 2021, pp. 8748--8763.

\bibitem{jia2021scaling}
C.~Jia, Y.~Yang, Y.~Xia, Y.-T. Chen, Z.~Parekh, H.~Pham, Q.~Le, Y.-H. Sung, Z.~Li, and T.~Duerig, ``Scaling up visual and vision-language representation learning with noisy text supervision,'' in \emph{ICML}.\hskip 1em plus 0.5em minus 0.4em\relax PMLR, 2021, pp. 4904--4916.

\bibitem{dosovitskiy2021image}
A.~Dosovitskiy, ``An image is worth 16x16 words: Transformers for image recognition at scale,'' \emph{arXiv preprint arXiv:2010.11929}, 2020.

\bibitem{li2021align}
J.~Li, R.~Selvaraju, A.~Gotmare, S.~Joty, C.~Xiong, and S.~C.~H. Hoi, ``Align before fuse: Vision and language representation learning with momentum distillation,'' \emph{NeurIPS}, 2021.

\bibitem{li2022blip}
J.~Li, D.~Li, C.~Xiong, and S.~Hoi, ``Blip: Bootstrapping language-image pre-training for unified vision-language understanding and generation,'' in \emph{ICML}.\hskip 1em plus 0.5em minus 0.4em\relax PMLR, 2022, pp. 12\,888--12\,900.

\bibitem{alayrac2022flamingo}
J.-B. Alayrac, J.~Donahue, P.~Luc, A.~Miech, I.~Barr, Y.~Hasson, K.~Lenc, A.~Mensch, K.~Millican, M.~Reynolds \emph{et~al.}, ``Flamingo: a visual language model for few-shot learning,'' \emph{NeurIPS}, vol.~35, pp. 23\,716--23\,736, 2022.

\bibitem{liu2023visual}
H.~Liu, C.~Li, Q.~Wu, and Y.~J. Lee, ``Visual instruction tuning,'' \emph{NeurIPS}, vol.~36, pp. 34\,892--34\,916, 2023.

\bibitem{yu2022coca}
J.~Yu, Z.~Wang, V.~Vasudevan, L.~Yeung, M.~Seyedhosseini, and Y.~Wu, ``Coca: Contrastive captioners are image-text foundation models,'' \emph{arXiv preprint arXiv:2205.01917}, 2022.

\bibitem{zhai2023sigmoid}
X.~Zhai, B.~Mustafa, A.~Kolesnikov, and L.~Beyer, ``Sigmoid loss for language image pre-training,'' in \emph{ICCV}, 2023, pp. 11\,975--11\,986.

\bibitem{chen2024internvl}
Z.~Chen, J.~Wu, W.~Wang, W.~Su, G.~Chen, S.~Xing, M.~Zhong, Q.~Zhang, X.~Zhu, L.~Lu \emph{et~al.}, ``Internvl: Scaling up vision foundation models and aligning for generic visual-linguistic tasks,'' in \emph{CVPR}, 2024, pp. 24\,185--24\,198.

\bibitem{zhu2024languagebind}
B.~Zhu, B.~Lin, M.~Ning, Y.~Yan, J.~Cui, H.~Wang, Y.~Pang, W.~Jiang, J.~Zhang, Z.~Li \emph{et~al.}, ``Languagebind: Extending video-language pretraining to n-modality by language-based semantic alignment,'' \emph{arXiv preprint arXiv:2310.01852}, 2024.

\bibitem{tu2023unicorns}
H.~Tu, C.~Cui, Z.~Wang, Y.~Zhou, B.~Zhao, J.~Han, W.~Zhou, H.~Yao, and C.~Xie, ``How many unicorns are in this image? a safety evaluation benchmark for vision llms,'' \emph{arXiv preprint arXiv:2311.16101}, 2023.

\bibitem{vatsa2023adventures}
M.~Vatsa, A.~Jain, and R.~Singh, ``Adventures of trustworthy vision-language models: A survey,'' \emph{arXiv preprint arXiv:2312.04231}, 2023.

\bibitem{liu2024safety}
X.~Liu, Y.~Zhu, Y.~Lan, C.~Yang, and Y.~Qiao, ``Safety of multimodal large language models on images and text,'' \emph{arXiv preprint arXiv:2402.00357}, 2024.

\bibitem{schlarmann2023adversarial}
C.~Schlarmann and M.~Hein, ``On the adversarial robustness of multi-modal foundation models,'' in \emph{ICCV}, 2023, pp. 3677--3685.

\bibitem{wei2023adaptive}
Z.~Wei, J.~Chen, Z.~Wu, and Y.-G. Jiang, ``Adaptive cross-modal transferable adversarial attacks from images to videos,'' \emph{TPAMI}, vol.~46, no.~5, pp. 3772--3783, 2023.

\bibitem{wei2023unified}
X.~Wei, Y.~Huang, Y.~Sun, and J.~Yu, ``Unified adversarial patch for visible-infrared cross-modal attacks in the physical world,'' \emph{TPAMI}, vol.~46, no.~4, pp. 2348--2363, 2023.

\bibitem{li2025attack}
F.~Li, T.~Wang, L.~Zhu, J.~Li, and H.~T. Shen, ``Attack as defense: Proactive adversarial multi-modal learning to evade retrieval,'' \emph{TPAMI}, 2025.

\bibitem{madry2019}
A.~Madry, A.~Makelov, L.~Schmidt, D.~Tsipras, and A.~Vladu, ``Towards deep learning models resistant to adversarial attacks,'' \emph{arXiv preprint arXiv:1706.06083}, 2019.

\bibitem{li-etal-2020-bert-attack}
\BIBentryALTinterwordspacing
L.~Li, R.~Ma, Q.~Guo, X.~Xue, and X.~Qiu, ``{BERT}-{ATTACK}: Adversarial attack against {BERT} using {BERT},'' in \emph{EMNLP}, B.~Webber, T.~Cohn, Y.~He, and Y.~Liu, Eds.\hskip 1em plus 0.5em minus 0.4em\relax Online: ACL, Nov. 2020, pp. 6193--6202. [Online]. Available: \url{https://aclanthology.org/2020.emnlp-main.500/}
\BIBentrySTDinterwordspacing

\bibitem{zhang2022adversariala}
J.~Zhang, Q.~Yi, and J.~Sang, ``Towards adversarial attack on vision-language pre-training models,'' in \emph{ACM MM}, ser. MM '22.\hskip 1em plus 0.5em minus 0.4em\relax New York, NY, USA: Association for Computing Machinery, Oct. 2022, pp. 5005--5013.

\bibitem{lu2023setlevel}
D.~Lu, Z.~Wang, T.~Wang, W.~Guan, H.~Gao, and F.~Zheng, ``Set-level guidance attack: Boosting adversarial transferability of vision-language pre-training models,'' in \emph{ICCV}, 2023, pp. 102--111.

\bibitem{fu2024improvingadversarialtransferabilityvisionlanguage}
J.~Fu, Z.~Chen, K.~Jiang, H.~Guo, J.~Wang, S.~Gao, and W.~Zhang, ``Improving adversarial transferability of vision-language pre-training models through collaborative multimodal interaction,'' \emph{arXiv preprint arXiv:2403.10883}, 2024.

\bibitem{qi2024visual}
X.~Qi, K.~Huang, A.~Panda, P.~Henderson, M.~Wang, and P.~Mittal, ``Visual adversarial examples jailbreak aligned large language models,'' in \emph{AAAI}, vol.~38, no.~19, 2024, pp. 21\,527--21\,536.

\bibitem{gong2024figstep}
Y.~Gong, D.~Ran, J.~Liu, C.~Wang, T.~Cong, A.~Wang, S.~Duan, and X.~Wang, ``Figstep: Jailbreaking large vision-language models via typographic visual prompts,'' in \emph{AAAI}, vol.~39, no.~22, 2025, pp. 23\,951--23\,959.

\bibitem{bailey2024image}
L.~Bailey, E.~Ong, S.~Russell, and S.~Emmons, ``Image hijacks: Adversarial images can control generative models at runtime,'' \emph{arXiv preprint arXiv:2309.00236}, 2023.

\bibitem{shayegani2024jailbreak}
E.~Shayegani, Y.~Dong, and N.~Abu-Ghazaleh, ``Jailbreak in pieces: Compositional adversarial attacks on multi-modal language models,'' \emph{arXiv preprint arXiv:2307.14539}, 2023.

\bibitem{xie2025chain}
P.~Xie, Y.~Bie, J.~Mao, Y.~Song, Y.~Wang, H.~Chen, and K.~Chen, ``Chain of attack: On the robustness of vision-language models against transfer-based adversarial attacks,'' in \emph{CVPR}, 2025.

\bibitem{zhang2025anyattack}
J.~Zhang, J.~Ye, X.~Ma, Y.~Li, Y.~Yang, Y.~Chen, J.~Sang, and D.-Y. Yeung, ``Anyattack: Towards large-scale self-supervised adversarial attacks on vision-language models,'' in \emph{CVPR}, 2025, pp. 19\,900--19\,909.

\bibitem{carlini2023aligned}
N.~Carlini, M.~Nasr, C.~A. Choquette-Choo, M.~Jagielski, I.~Gao, P.~W.~W. Koh, D.~Ippolito, F.~Tramer, and L.~Schmidt, ``Are aligned neural networks adversarially aligned?'' \emph{NeurIPS}, vol.~36, pp. 61\,478--61\,500, 2023.

\bibitem{gu2023survey}
J.~Gu, X.~Jia, P.~de~Jorge, W.~Yu, X.~Liu, A.~Ma, Y.~Xun, A.~Hu, A.~Khakzar, Z.~Li \emph{et~al.}, ``A survey on transferability of adversarial examples across deep neural networks,'' \emph{ICCV}, 2023.

\bibitem{dong2018boosting}
Y.~Dong, F.~Liao, T.~Pang, H.~Su, J.~Zhu, X.~Hu, and J.~Li, ``Boosting adversarial attacks with momentum,'' in \emph{CVPR}, 2018.

\bibitem{liu2016delving}
Y.~Liu, X.~Chen, C.~Liu, and D.~Song, ``Delving into transferable adversarial examples and black-box attacks,'' \emph{ICLR}, 2017.

\bibitem{xie2019improving}
C.~Xie, Z.~Zhang, Y.~Zhou, S.~Bai, J.~Wang, Z.~Ren, and A.~L. Yuille, ``Improving transferability of adversarial examples with input diversity,'' in \emph{CVPR}, 2019.

\bibitem{dong2019evading}
Y.~Dong, T.~Pang, H.~Su, and J.~Zhu, ``Evading defenses to transferable adversarial examples by translation-invariant attacks,'' in \emph{CVPR}, 2019.

\bibitem{athalye2018synthesizing}
A.~Athalye, L.~Engstrom, A.~Ilyas, and K.~Kwok, ``Synthesizing robust adversarial examples,'' in \emph{ICML}, 2018.

\bibitem{brown2017adversarial}
T.~B. Brown, D.~Man{\'e}, A.~Roy, M.~Abadi, and J.~Gilmer, ``Adversarial patch,'' \emph{arXiv preprint arXiv:1712.09665}, 2017.

\bibitem{eykholt2018robust}
K.~Eykholt, I.~Evtimov, E.~Fernandes, B.~Li, A.~Rahmati, C.~Xiao, A.~Prakash, T.~Kohno, and D.~Song, ``Robust physical-world attacks on deep learning visual classification,'' in \emph{CVPR}, 2018.

\bibitem{moosavi2017universal}
S.-M. Moosavi-Dezfooli, A.~Fawzi, O.~Fawzi, and P.~Frossard, ``Universal adversarial perturbations,'' in \emph{CVPR}, 2017.

\bibitem{zhao2024survey}
T.~Zhao, L.~Zhang, Y.~Ma, and L.~Cheng, ``A survey on safe multi-modal learning system,'' \emph{arXiv preprint arXiv:2402.05355}, 2024.

\bibitem{liu2024survey}
D.~Liu, M.~Yang, X.~Qu, P.~Zhou, Y.~Cheng, and W.~Hu, ``A survey of attacks on large vision-language models: Resources, advances, and future trends,'' \emph{arXiv preprint arXiv:2407.07403}, 2024.

\bibitem{li2024onea}
L.~Li, H.~Guan, J.~Qiu, and M.~Spratling, ``One prompt word is enough to boost adversarial robustness for pre-trained vision-language models,'' in \emph{CVPR}, 2024, pp. 24\,408--24\,419.

\bibitem{zhang2024adversariala}
J.~Zhang, X.~Ma, X.~Wang, L.~Qiu, J.~Wang, Y.-G. Jiang, and J.~Sang, ``Adversarial prompt tuning for vision-language models,'' \emph{arXiv preprint arXiv:2311.11261}, Aug. 2024.

\bibitem{mao2023understanding}
C.~Mao, S.~Geng, J.~Yang, X.~Wang, and C.~Vondrick, ``Understanding zero-shot adversarial robustness for large-scale models,'' in \emph{ICLR}.\hskip 1em plus 0.5em minus 0.4em\relax arXiv, Apr. 2023.

\bibitem{wang2024pretraineda}
S.~Wang, J.~Zhang, Z.~Yuan, and S.~Shan, ``Pre-trained model guided fine-tuning for zero-shot adversarial robustness,'' in \emph{CVPR}, 2024, pp. 24\,502--24\,511.

\bibitem{schlarmann2024robusta}
C.~Schlarmann, N.~D. Singh, F.~Croce, and M.~Hein, ``Robust clip: Unsupervised adversarial fine-tuning of vision embeddings for robust large vision-language models,'' \emph{arXiv preprint arXiv:2402.12336}, Jun. 2024.

\bibitem{waseda2024leveraging}
F.~Waseda and A.~{Tejero-de-Pablos}, ``Leveraging many-to-many relationships for defending against visual-language adversarial attacks,'' \emph{arXiv preprint arXiv:2405.18770}, 2024.

\bibitem{zhou2024revisitinga}
W.~Zhou, S.~Bai, D.~P. Mandic, Q.~Zhao, and B.~Chen, ``Revisiting the adversarial robustness of vision language models: A multimodal perspective,'' \emph{arXiv preprint arXiv:2404.19287}, 2024.

\bibitem{tu2023closer}
W.~Tu, W.~Deng, and T.~Gedeon, ``A closer look at the robustness of contrastive language-image pre-training (clip),'' \emph{NeurIPS}, vol.~36, pp. 13\,678--13\,691, 2023.

\bibitem{tgazsr2024}
L.~Yu, H.~Zhang, and C.~Xu, ``Text-guided attention is all you need for zero-shot robustness in vision-language models,'' \emph{NeurIPS}, vol.~37, pp. 96\,424--96\,448, 2024.

\bibitem{robey2024smoothllm}
A.~Robey, E.~Wong, H.~Hassani, and G.~J. Pappas, ``Smoothllm: Defending large language models against jailbreaking attacks,'' \emph{arXiv preprint arXiv:2310.03684}, 2023.

\bibitem{liu2024mmsafetybench}
X.~Liu, Y.~Zhu, J.~Gu, Y.~Lan, C.~Yang, and Y.~Qiao, ``Mm-safetybench: A benchmark for safety evaluation of multimodal large language models,'' in \emph{ECCV}.\hskip 1em plus 0.5em minus 0.4em\relax Springer, 2024, pp. 386--403.

\bibitem{zhang2019theoretically}
H.~Zhang, Y.~Yu, J.~Jiao, E.~Xing, L.~E. Ghaoui, and M.~Jordan, ``Theoretically principled trade-off between robustness and accuracy,'' in \emph{ICML}.\hskip 1em plus 0.5em minus 0.4em\relax PMLR, May 2019, pp. 7472--7482.

\bibitem{raghunathan2019adversarial}
A.~Raghunathan and e.~a. Xie, Sang~Michael, ``Adversarial training can hurt generalization,'' in \emph{ICML Workshop}, May 2019.

\bibitem{liang2024mind}
W.~Liang, Y.~Zhang, Y.~Kwon, S.~Yeung, and J.~Zou, ``Mind the gap: Understanding the modality gap in multi-modal contrastive representation learning,'' in \emph{NeurIPS}, ser. NeurIPS.\hskip 1em plus 0.5em minus 0.4em\relax Red Hook, NY, USA: Curran Associates Inc., Apr. 2022, pp. 17\,612--17\,625.

\bibitem{deng2009imagenet}
J.~Deng, W.~Dong, R.~Socher, L.-J. Li, K.~Li, and L.~{Fei-Fei}, ``Imagenet: A large-scale hierarchical image database,'' in \emph{CVPR}, 2009.

\bibitem{peng2018overview}
Y.~Peng, X.~Huang, and Y.~Zhao, ``An overview of cross-media retrieval: Concepts, methodologies, benchmarks, and challenges,'' \emph{IEEE Transactions on Circuits and Systems for Video Technology}, vol.~28, no.~9, pp. 2372--2385, Sep. 2018.

\bibitem{lin2014microsoft}
T.-Y. Lin, M.~Maire, S.~Belongie, J.~Hays, P.~Perona, D.~Ramanan, P.~Doll{\'a}r, and C.~L. Zitnick, ``Microsoft coco: Common objects in context,'' in \emph{ECCV}, 2014, pp. 740--755.

\bibitem{andriushchenko2020square}
M.~Andriushchenko, F.~Croce, N.~Flammarion, and M.~Hein, ``Square attack: A query-efficient black-box adversarial attack via random search,'' in \emph{ECCV}, 2020, pp. 484--501.

\bibitem{ilharco_gabriel_2021_5143773}
\BIBentryALTinterwordspacing
G.~Ilharco, M.~Wortsman, R.~Wightman, C.~Gordon, N.~Carlini, R.~Taori, A.~Dave, V.~Shankar, H.~Namkoong, J.~Miller, H.~Hajishirzi, A.~Farhadi, and L.~Schmidt, ``Openclip,'' Jul. 2021, if you use this software, please cite it as below. [Online]. Available: \url{https://doi.org/10.5281/zenodo.5143773}
\BIBentrySTDinterwordspacing

\bibitem{guzhov2022audioclip}
A.~Guzhov, F.~Raue, J.~Hees, and A.~Dengel, ``Audioclip: Extending clip to image, text and audio,'' in \emph{ICASSP}, 2022, pp. 976--980.

\bibitem{gemmeke2017audio}
J.~F. Gemmeke, D.~P. Ellis, D.~Freedman, A.~Jansen, W.~Lawrence, R.~C. Moore, M.~Plakal, and M.~Ritter, ``Audio set: An ontology and human-labeled dataset for audio events,'' in \emph{ICASSP}, 2017.

\bibitem{kim2019audiocaps}
C.~D. Kim, B.~Kim, H.~Lee, and G.~Kim, ``Audiocaps: Generating captions for audios in the wild,'' in \emph{NAACL}, 2019.

\bibitem{shi2020differentiable}
J.~Shi, E.~Riba, D.~Mishkin, F.~Moreno, and A.~Nicolaou, ``Differentiable data augmentation with kornia,'' \emph{arXiv preprint arXiv:2011.09832}, 2020.

\bibitem{nie2022diffusion}
W.~Nie, B.~Guo, Y.~Huang, C.~Xiao, A.~Vahdat, and A.~Anandkumar, ``Diffusion models for adversarial purification,'' \emph{ICML}, 2022.

\bibitem{hill2020stochastic}
M.~Hill, J.~Mitchell, and S.-C. Zhu, ``Stochastic security: Adversarial defense using long-run dynamics of energy-based models,'' \emph{ICLR}, 2021.

\bibitem{roth2023wafflinga}
K.~Roth, J.~M. Kim, A.~Sophia~Koepke, O.~Vinyals, C.~Schmid, and Z.~Akata, ``Waffling around for performance: Visual classification with random words and broad concepts,'' in \emph{ICCV}, Oct. 2023, pp. 15\,700--15\,711.

\end{thebibliography}
%

\vspace{-20mm}
\begin{IEEEbiography}[{\includegraphics[width=1in,height=1.25in,clip,keepaspectratio]{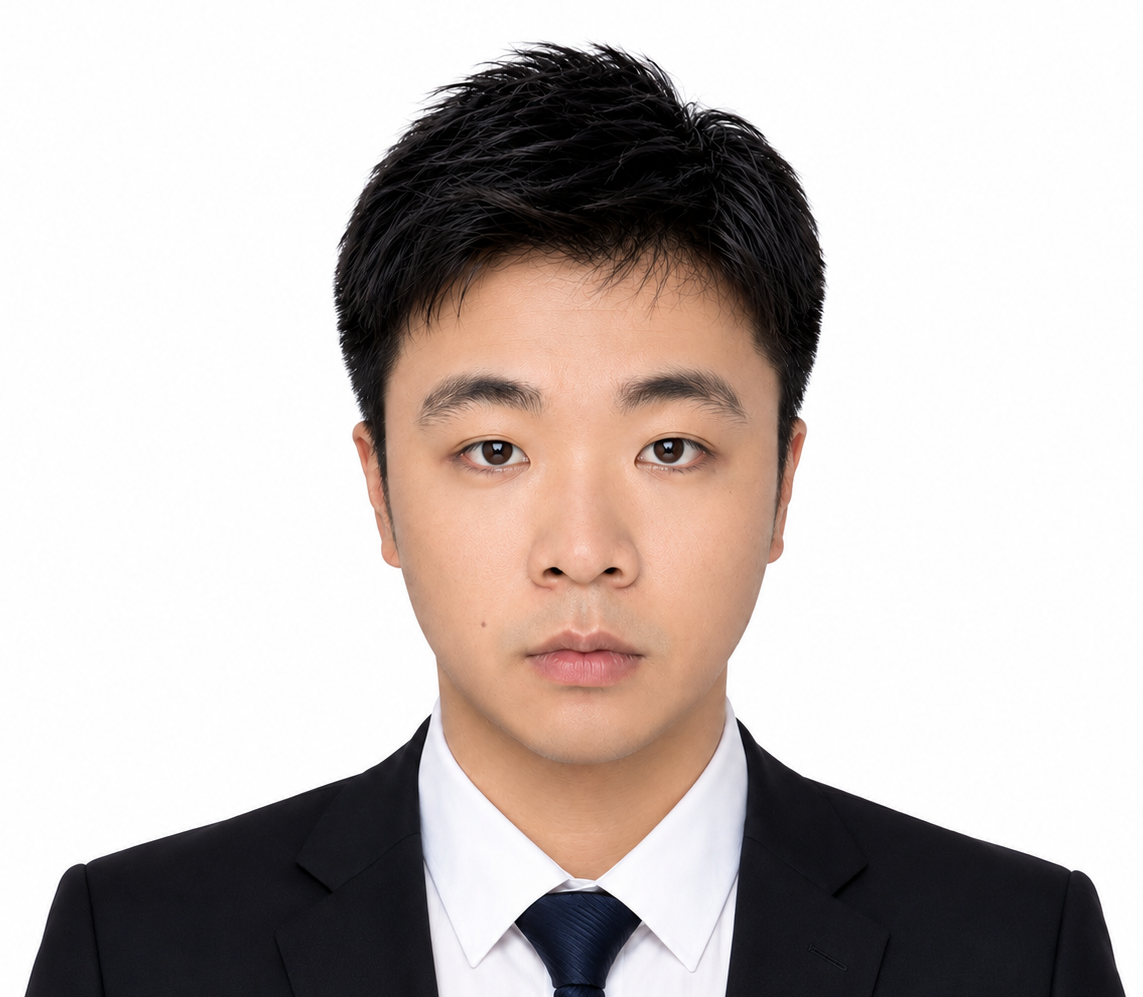}}]{Zhifang Zhang} is a first-year PhD student at the University of Queensland, Australia under the co-supervision of Dr. Miao Xu and Dr. Lei Feng. 
His research focuses on AI Security and Multimodal Models.
He regularly serves as a reviewer for ICLR, CVPR, etc.
\vspace{-10mm}
\end{IEEEbiography}
\begin{IEEEbiography}[{\includegraphics[width=1in,height=1.25in,clip,keepaspectratio]{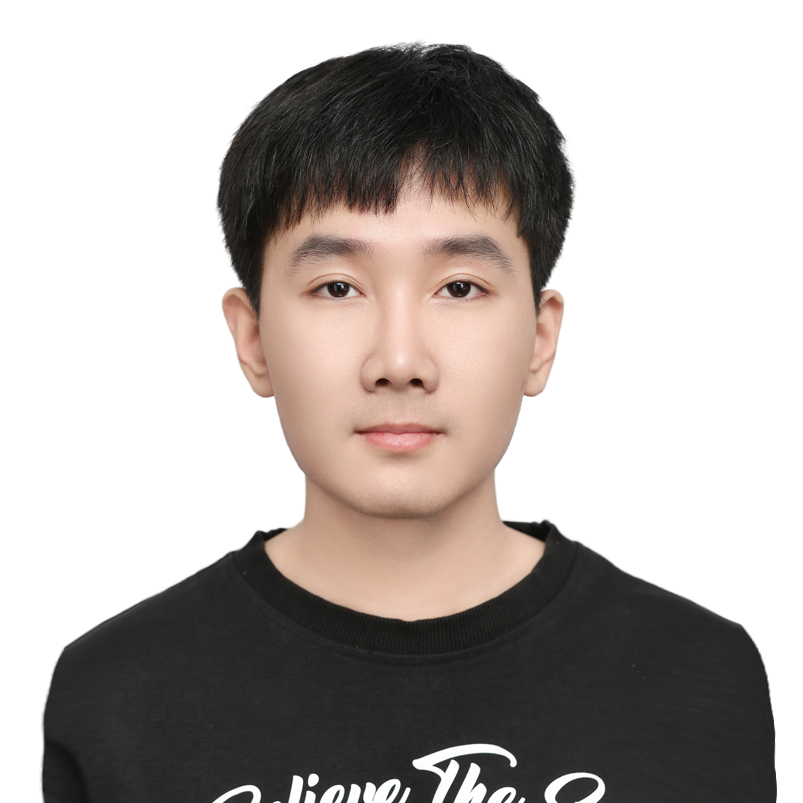}}]{Jiahan Zhang} is a first-year PhD student at Johns Hopkins University, USA under the supervision of Dr. Jieneng Chen.
His research focuses on World Models and AI Security.
He regularly serves as a reviewer of top conferences such as ICLR, NeurIPS and ICML.
\vspace{-10mm}
\end{IEEEbiography}
\begin{IEEEbiography}[{\includegraphics[width=1in,height=1.25in,clip,keepaspectratio]{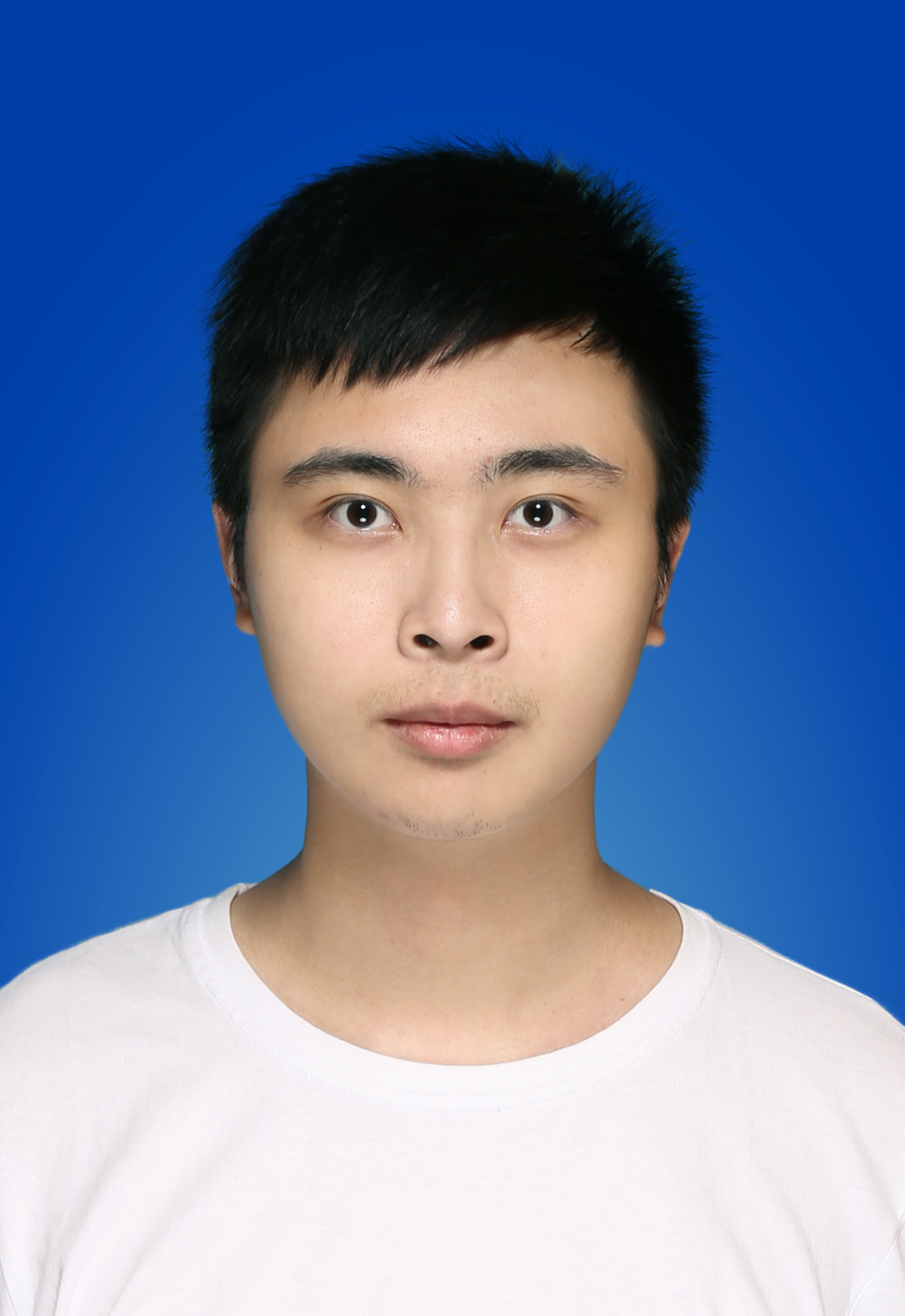}}]{Shengjie Zhou} is a master student at Chongqing University, China under the supervision of Dr. Lei Feng. 
His research focuses on Adversarial Attacks and Defenses.
He regularly serves as a reviewer for top conferences or journals such as ICLR and PR.
\vspace{-10mm}
\end{IEEEbiography}
\begin{IEEEbiography}[{\includegraphics[width=1in,height=1.25in,clip,keepaspectratio]{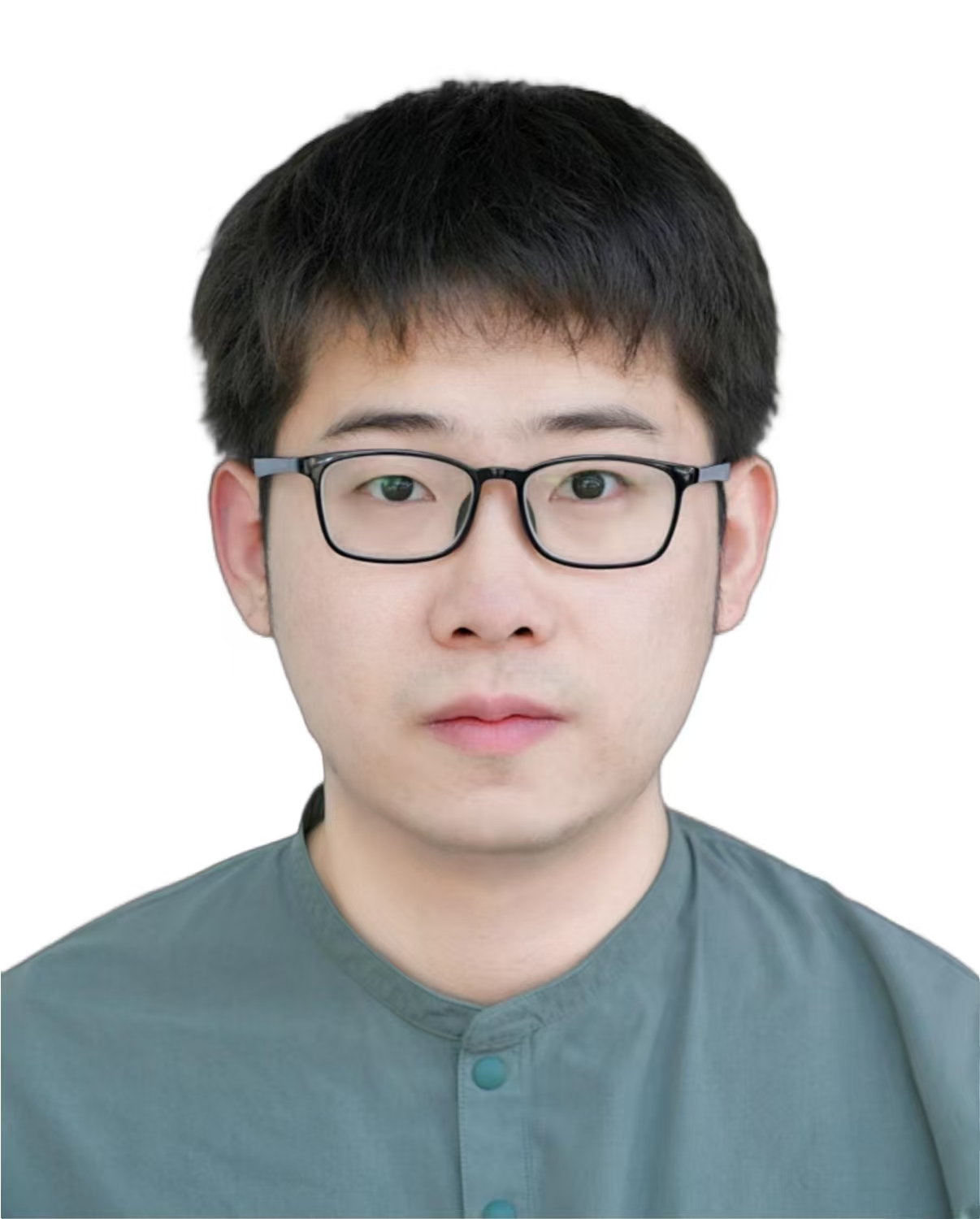}}]{Shuo He} is a Postdoctoral Research Fellow at Nanyang Technological University. His research focuses on AI safety and LLM-based Agent.
He received his Ph.D. degree from University of Electronic Science and Technology of China in 2024.
Shuo He regularly serves as a reviewer for top conferences such as ICLR, ICML and NeurIPS, and top journals such as PR and TPAMI.
\vspace{-10mm}
\end{IEEEbiography}
\begin{IEEEbiography}[{\includegraphics[width=1in,height=1.25in,clip,keepaspectratio]{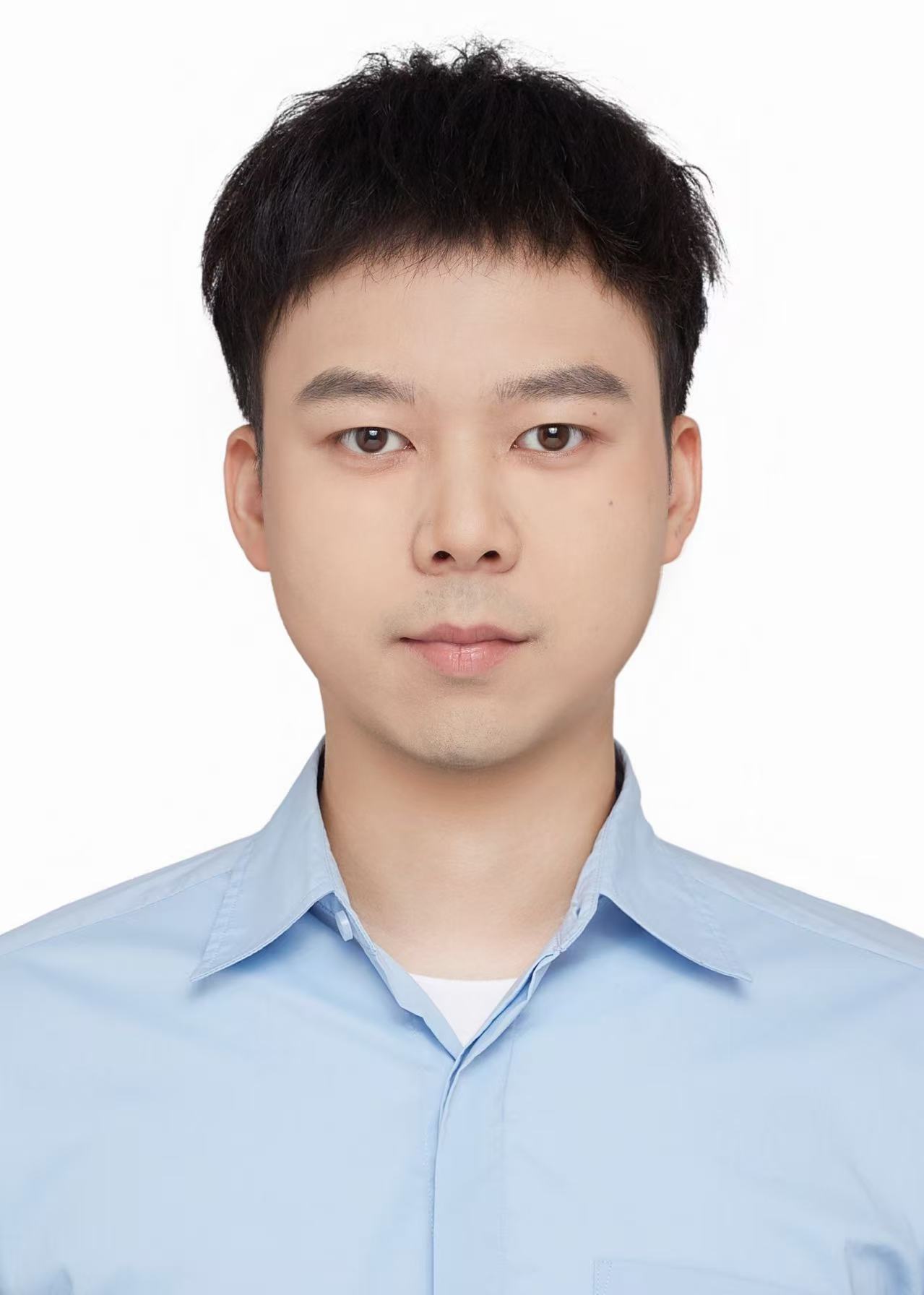}}]{Qi Wei} is a second-year Ph.D. candidate at Nanyang Technological University. 
His research focuses on include computer vision and machine learning, such as AI safety and Weakly-supervised Learning.
Qi Wei regularly serves as a reviewer for top conferences such as ICLR, ICML and NeurIPS, and top journals such as MLJ and TNNLS.
\vspace{-10mm}
\end{IEEEbiography}
\begin{IEEEbiography}[{\includegraphics[width=1in,height=1.25in,clip,keepaspectratio]{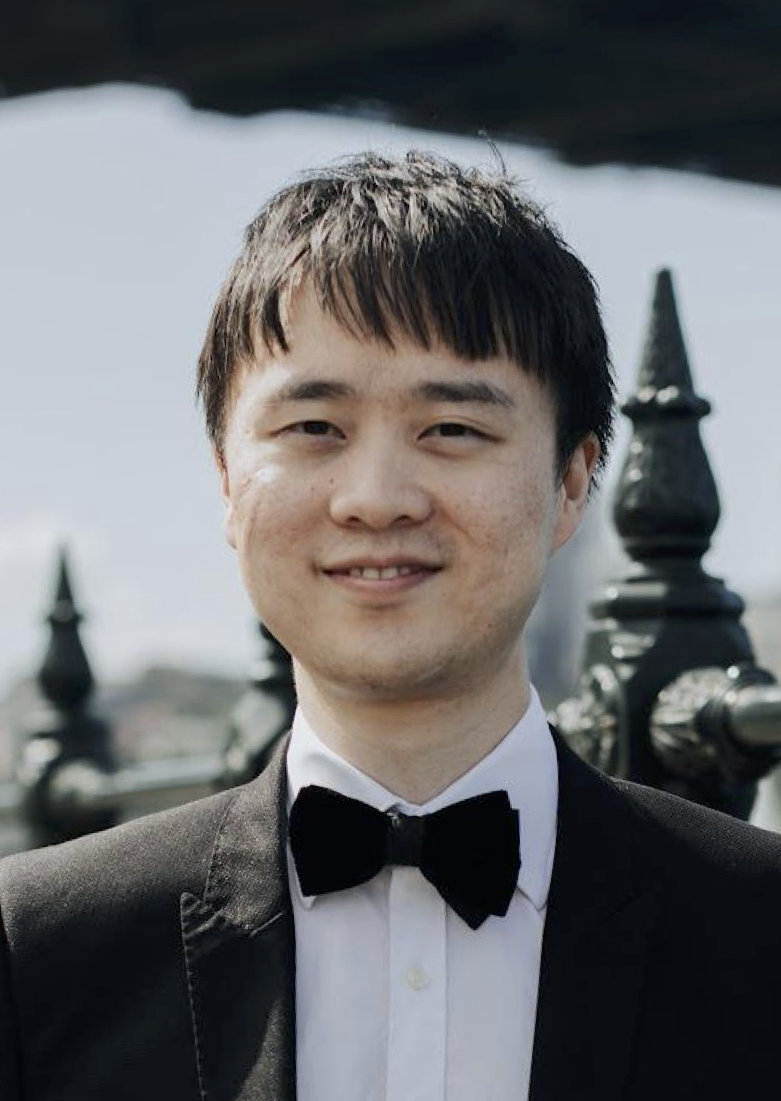}}]{Feng Liu} received the BSc degree in mathematics and the MSc degree in probability and statistics from the School of Mathematics and Statistics, Lanzhou University, Lanzhou, China, in 2013 and 2015, respectively, and the PhD degree in computer science from the University of Technology Sydney, Sydney, NSW, Australia, in 2020. He is currently a lecturer with the School of Computing and Information Systems, The University of Melbourne, Melbourne, VIC, Australia. His research interests include hypothesis testing and trustworthy machine learning. Dr. Liu has served as an area chair for NeurIPS, ICML, ICLR, AISTATS and ACMMM. He also served as an editor for ACM TOPML, associate editor for IJMLC, action editor for Neural Networks, action editor for Transactions on Machine Learning Research and reviewers for many academic journals, such as Journal of Machine Learning Research, IEEE Transactions on Pattern Analysis and Machine Intelligence, Transactions on Machine Learning Research, MLJ, and so on. He rreceived the ARC Discovery Early Career Researcher Award and the Outstanding Paper Award of NeurIPS (2022).
\vspace{-10mm}
\end{IEEEbiography}

\begin{IEEEbiography}[{\includegraphics[width=1in,height=1.25in,clip,keepaspectratio]{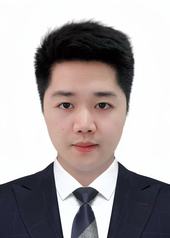}}] {Lei Feng} received his Ph.D. degree from Nanyang Technological University, Singapore, in 2021. He is currently a Full Professor at the School of Computer Science and Engineering, Southeast University, China. His main research interests include AI agents and multimodal large language models. He has published more than 100 papers at premier conferences and journals. He has regularly served as an Area Chair for ICML, NeurIPS, and ICLR, and served as an Action Editor for Neural Networks and Transactions on Machine Learning Research. He has received the ICLR 2022 Outstanding Paper Award Honorable Mention and the WAIC 2024 Youth Outstanding Paper Nomination Award. He was named to Forbes 30 Under 30 China 2021, Forbes 30 Under 30 Asia 2022, and Asia-Pacific Leaders Under 30 in 2024.
\vspace{-10mm}
\end{IEEEbiography}
 



	
	 
\end{document}